\pdfoutput=1
\documentclass[letterpaper,twocolumn,10pt]{article}
\usepackage{usenix}
\usepackage{authblk}
\usepackage{amsmath}
\usepackage{amssymb}
\usepackage{filecontents}
\usepackage{amsthm}
\newtheorem{theorem}{Theorem}
\newtheorem{lemma}{Lemma}
\usepackage{algorithm2e}
\SetKwInput{KwInput}{Input}                
\SetKwInput{KwOutput}{Output}              
\usepackage{graphicx}
\usepackage{tablefootnote}
\usepackage{float}
\usepackage{tabularx}
\usepackage{caption}
\usepackage{subcaption}
\usepackage{threeparttable}

\usepackage{tcolorbox}
\usepackage[absolute]{textpos}

\usepackage{booktabs}
\usepackage{multirow}
\usepackage{diagbox}
\usepackage{arydshln}
\usepackage{makecell}

\begin{document}
\begin{textblock}{16}(3,1.5)
To appear in the 32nd USENIX Security Symposium, August 2023, Anaheim, CA, USA
\end{textblock}

\date{}

\title{\Large \bf Inductive Graph Unlearning}

\author[1,4]{\rm Cheng-Long Wang}
\author[2]{\rm Mengdi Huai}
\author[1,3,4]{\rm Di Wang}
\affil[1]{King Abdullah University of Science and Technology}
\affil[2]{Iowa State University}
\affil[3]{Computational Bioscience Research Center}
\affil[4]{SDAIA-KAUST Center of Excellence in Data Science and Artificial Intelligence}



\maketitle

\begin{abstract}
As a way to implement the "right to be forgotten" in machine learning, \textit{machine unlearning} aims to completely remove the contributions and information of the samples to be deleted from a trained model without affecting the contributions of other samples.  Recently, many frameworks for machine unlearning have been proposed, and most of them focus on image and text data. To extend machine unlearning to graph data, \textit{GraphEraser} has been proposed. However, a critical issue is that \textit{GraphEraser} is specifically designed for the transductive graph setting, where the graph is static and attributes and edges of test nodes are visible during training. It is unsuitable for the inductive setting, where the graph could be dynamic and the test graph information is invisible in advance. Such inductive capability is essential for production machine learning systems with evolving graphs like social media and transaction networks. To fill this gap, we propose the \underline{{\bf G}}\underline{{\bf U}}ided \underline{{\bf I}}n\underline{{\bf D}}uctiv\underline{{\bf E}} Graph Unlearning framework (GUIDE). GUIDE consists of three components: guided graph partitioning with fairness and balance, efficient subgraph repair, and similarity-based aggregation. Empirically, we evaluate our method on several inductive benchmarks and evolving transaction graphs. Generally speaking, GUIDE can be efficiently implemented on the inductive graph learning tasks for its low graph partition cost, no matter on computation or structure information. The code will be available here: https://github.com/Happy2Git/GUIDE.
\end{abstract}

\section{Introduction}
 In various complex real-world applications, we often encounter cases where the data is represented as graphs, such as medical diagnosis\cite{JinSEIZCJB21}, social media\cite{QiuTMDW018}, advertising industry\cite{YingHCEHL18}, and financial industry\cite{SukharevSFPB20}. The interactions between neighboring nodes make it promising to learn rich information from graph data. After showing great promise in effectively solving graph-based machine learning tasks such as node classification, link prediction, and graph classification, Graph Neural Networks (GNNs) with their large number of variants \cite{KipfW17,HamiltonYL17,XuHLJ19,WuSZFYW19} have received much attention from the machine learning community. Despite their success, recent deployments of GNNs simultaneously raise privacy concerns when the input graphs contain sensitive information of personal data, such as social networks and biomedical data.  Recently, the "right to be forgotten" has been proposed in many regulations to protect users' personal information, such as the European Union's General Data Protection Regulation (GDPR) and the California Consumer Privacy Act (CCPA) ~\cite{rosen2011right,politou2018backups,pardau2018california,RuanXJWSH21}. Broadly speaking, the "right to be forgotten" provides individuals the right to request the deletion of their personal information and the right to opt out of the sale of their personal information.

\begin{figure*}[t]
\centering
\includegraphics[width=0.95\textwidth,scale=0.5]{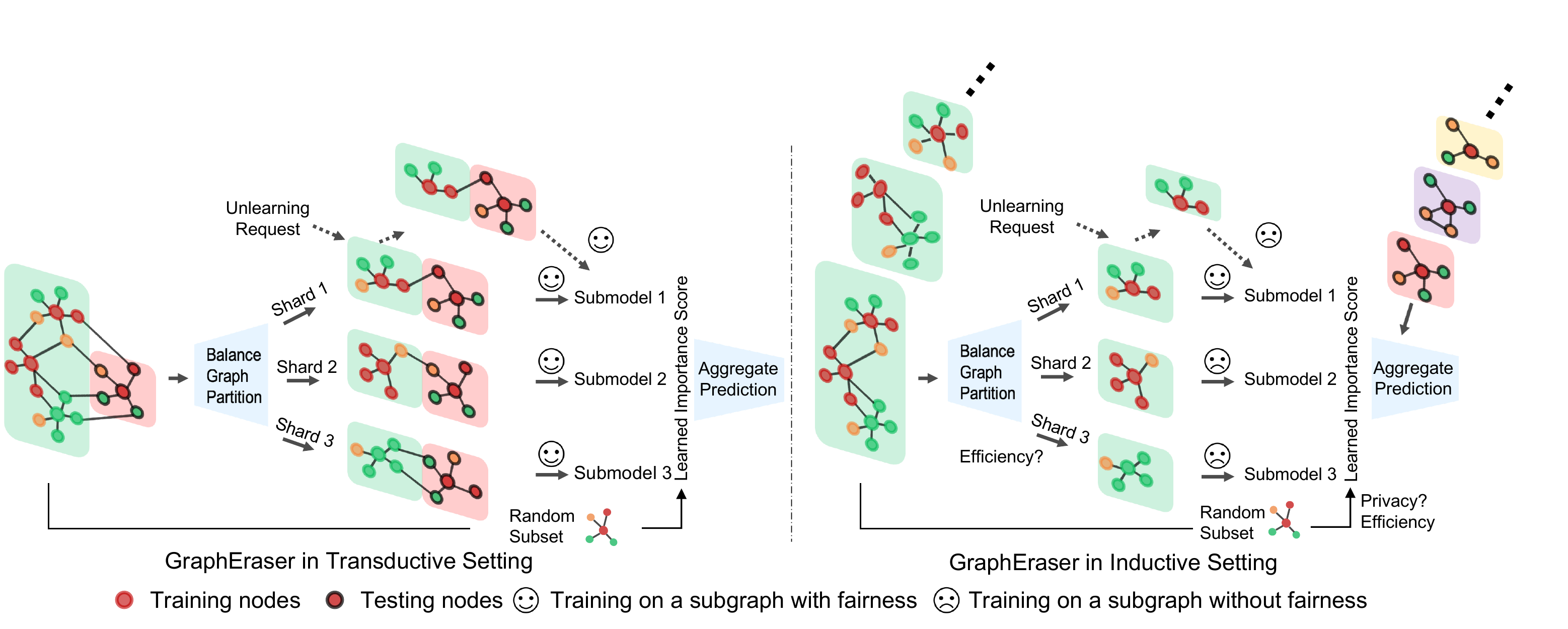}
\caption{Behavior of GraphEraser in various settings. The colors represent the ground truth labels of the corresponding nodes. In the transductive setting, the features of the test nodes and their connections to other nodes are visible during the training process (in the same static graph with training nodes). The red-shaded subgraphs indicate the test nodes (surrounded by black circles) whose labels are unknown to the model owner in advance. In the inductive setting, training graphs can evolve over time or change incrementally. The test graphs are also completely invisible, resulting in limited information available for training each shard (with a small subgraph) without access to the test nodes and their edges.}
\label{Picture1}
\vspace{-0.15in}
\end{figure*}

As a de facto way to implement the "right to be forgotten" in machine learning, machine unlearning allows the model owner to completely remove the trace of the samples to be deleted from a trained model without affecting the contributions of other samples, while its unlearning process requires significantly lower computational cost than retraining from scratch. In recent years, a long list of work on machine unlearning has been proposed, and these methods can be categorized into two classes: model-agnostic unlearning~\cite{CaoY15,WuDD20,BourtouleCCJTZL21,GuptaJNRSW21} and model-intrinsic unlearning~\cite{GinartGVZ19,BrophyL21,GolatkarARPS21,IzzoSCZ21,SekhariAKS21,ThudiDCP22}. As one of the most well-known model-agnostic methods, SISA~\cite{BourtouleCCJTZL21} uses data partitioning mechanisms to achieve efficient unlearning without full retraining. Specifically, it first divides the dataset into multiple isolated shards and trains a submodel for each shard.  Then it aggregates the predictions of all submodels to obtain the final prediction. Such submodels can limit the influence of each data sample throughout the training process. When there is a data removal request, the model owner only needs to partially retrain the submodel corresponding to the data to be removed. Compared to SISA, many unlearning methods, instead of retraining submodels, aim to obtain a shifted model that satisfies some unlearning criteria by modifying the weights of the existing trained model~\cite{ThudiDCP22}. While these unlearning methods have computational advantages, they are not as transparent as SISA.

Although there are numerous studies on machine unlearning, most are only tailored for image or text data, and unlearning methods for graph data, i.e., graph unlearning, are still lacking. Due to the additional node dependency in graph data, existing unlearning methods cannot be directly applied, which indicates that graph unlearning is more challenging. Based on the SISA framework, \cite{abs-2103-14991} proposes the first graph unlearning method, GraphEraser, for graph neural network (GNN) models. Compared with the random partitioning in SISA, GraphEraser provides two balanced partition methods to preserve the additional structural information in graph data. Then, it applies a learning-based aggregation method to obtain the importance scores of submodels. Later, \cite{abs-2206-09140} proposes a Certified Graph Unlearning (CGU) method based on the Simplifying Graph Convolutional Network (SGC)~\cite{WuSZFYW19}, which is a linear GCN model. Unfortunately, such a model-specific method is inapplicable to general GNN models. 

However, as we will show later, GraphEraser and CGU are inherently designed for the \textit{transductive} graph setting, where the attributes (but not labels) and edges of test nodes are visible during training. They are not designed for the \textit{inductive} graph setting (where test nodes and edges are invisible during training), which is ubiquitous in high-throughput production machine learning systems, as pointed out by~\cite{HamiltonYL17}. For example, the evolving graphs in a transaction system constantly encounter unseen data samples every day. Thus, the associated fraud detection models should be able to generalize to the newly generated graphs efficiently.  Besides transaction systems, such inductive capabilities are also crucial for GNN models in social media, advertising, etc.

For GraphEraser, the time cost of graph partitioning is exceedingly high, so it is not suitable to implement this framework for the evolving graph or multi-graph cases in the inductive setting. Graph unlearning requires that each shard retains a small piece of the training graph to train a submodel. However, the loss of visibility of test nodes and their connections makes submodel training more difficult in the inductive setting. For example, it is easy to learn a weak submodel due to the unfair label composition in each shard, as shown in Figure~\ref{Picture1}.  Note that the \textit{fairness} here refers to group fairness, which ensures some form of statistical parity for members of different protected groups (e.g., gender or race)~\cite{FairnessNotion}, i.e., the label distribution in each shard remains the same statistic as in the entire training graph. And we use \textit{balance} in the following discussions to represent that the subgraph of each shard has the same size (number of nodes). In addition, GraphEraser aggregates the predictions of submodels on the test nodes by learning important scores for all shards. Once one shard is updated, all other shards need to retrain their important scores, which brings more computational cost and privacy risk.

Thus, we can conclude that the main challenge in model-agnostic inductive graph unlearning is to preserve as much structural information of the original graph as possible while satisfying both fairness and balance constraints in graph partitioning efficiently. This is based on the insights that more structural information leads to higher model performance, a balanced partition ensures that the expected unlearning time cost is small when facing small batch unlearning, and a fair partition would lead to a more robust learning process. 

\noindent {\bf Our contributions:} Motivated by our above findings, in this paper we propose the first inductive graph unlearning framework called GUided InDuctivE Graph Unlearning (GUIDE). Briefly, GUIDE consists of three components: guided graph partitioning with fairness and balance, efficient subgraph repairing, and similarity-based aggregation. Specifically, in guided graph partitioning, we propose two novel graph partitioning methods: GPFB-Fast and GPFB-SR, to obtain a graph partition that efficiently satisfies both fairness and balance constraints. According to our experimental results, the proposed methods are superior to GraphEraser with $\sim 3\times$ balance and fairness scores. GPFB-Fast achieves $\sim 10\times$ speedup on graph partitioning. To the best of our knowledge, this is also the first study on graph partitioning with fairness and balance constraints. 

Due to graph partitioning, a lot of edges would be lost, destroying the structure of the original graph. Therefore, to restore this missing information as much as possible, we propose subgraph repair methods as the second component of GUIDE. Through our methods, missing neighbors and their connections with the corresponding nodes could be efficiently generated and added to these subgraphs to repair their structure. Notably, for each shard, our repairing procedures do not involve the information of other shards. After receiving node removal requests, the corresponding repaired subgraphs can be efficiently updated by deleting the corresponding nodes and edges. 

As mentioned above, the learning-based aggregation method LBAggr proposed by ~\cite{abs-2103-14991} requires access to the entire training graph when updating the importance scores of the corresponding shards. To speed up the training process, LBAggr is trained on a constructed public subset of the training graph. However, once a shard is updated, all importance scores of other shards need to be updated as well, which introduces additional computational cost. We develop a novel similarity-based aggregation method as our third component to address these issues. Unlike previous methods, our method can compute the importance score of each shard independently, and the normalized similarity score between the partitioned subgraph and the test graph can be directly used as the corresponding importance score. Such independent updating will be more efficient than GrapnEraser when the unlearning batch size is small.

We perform extensive experiments to demonstrate the performance of GUIDE in the inductive setting. GUIDE
achieves superior performance ($\sim 3\times$) than the existing state-of-the-art methods on popular node classification benchmarks and the fraud detection task on a real bitcoin dataset. We also introduce two metrics to evaluate the graph partitioning results: balance score and fairness score. Specifically, experimental results show that GUIDE has lower time cost than \textit{GraphEraser} while achieving higher fairness and balance scores in graph partitioning.  In addition, we perform extensive ablation studies to demonstrate the utility of other components of GUIDE. Ablation studies show that our proposed subgraph repair methods can significantly improve the performance of GNN models trained on subgraphs. Furthermore, similarity-based aggregation can achieve comparable results to learning-based aggregation.

\section{Preliminaries}
\subsection{Graph Neural Networks}

Given an undirected graph $\mathcal{G=(V,E)}$, where $\mathcal{V}$ is the set of nodes and $\mathcal{E}$ is the set of edges, a basic graph neural network (GNN) model attempts to learn a node representation for downstream tasks from the graph structure and any feature information we have. To train a GNN model, we always use the message passing framework. 
During each iteration, the GNN model updates the node embedding for each node $u \in \mathcal{V}$ by aggregating the information from $u$'s neighbors $\mathcal{N}(u)$. The $k$-th update process can be formulated as follows~\cite{2020Hamilton}:
\begin{equation*}
\begin{split}
h_{u}^{(k+1)} & = \text{UPDATE}^{(k)}\left(h_{u}^{(k)}, \text{AGGR}^{(k)}({h_{v}^{(k)}, \forall v\in \mathcal{N}(u)}) \right) \\
& = \text{UPDATE}^{(k)}\left(h_{u}^{(k)}, m^{(k)}_{ \mathcal{N}(u)} \right),
\end{split}
\end{equation*}
where $\text{UPDATE}$ and $\text{AGGR}$ are some differentiable functions and $m^{(k)}_{ \mathcal{N}(u)}$ is the aggregated 'message' from the neighbors of $u$. After $K$ iterations of message passing, we can obtain the final embedding for each node. These node embeddings can be used for node classification, graph classification, and relation prediction tasks. 

\noindent
\textbf{Transductive and Inductive Graph Learning.}  There are two settings for node classification tasks: the transductive setting and the inductive setting. In the transductive setting, the training nodes and test nodes are in the same static graph. Test nodes and their associated edges are involved in GNN's message passing updates, even though they are unlabeled and not used in the loss computation. In contrast, all test nodes and their edges are completely unobservable during training in the inductive setting. Besides, the training graph can also evolve over time. Compared to the transductive setting, the inductive setting is more common in production machine learning systems that operate on evolving graphs and constantly encounter unseen nodes, such as the daily user-video graphs generated on Youtube~\cite{HamiltonYL17}.

\subsection{Transductive Graph Unlearning}
{\bf Machine Unlearning.} Machine unlearning aims to fully eliminate any influence of the data to be deleted from a trained machine learning (ML) model. 
To implement machine unlearning, the most natural approach is to directly delete all the revoked samples and retrain the ML model from scratch by using the original training data without deleted samples. While retraining from scratch is easy to implement, its computation cost will be prohibitively large to make it efficient when both the model and the training data are large-scale.  Later on, several methods have been proposed to reduce the computation
overhead. See the related work section \ref{related_work} in Appendix for details. 

\noindent {\bf Graph Unlearning.} Graph unlearning refers to machine unlearning for graph data, and in this paper we will focus on GNN learning models. Compared to the standard machine unlearning, there are additional challenges in graph unlearning, e.g. the node dependency in graph data makes most of the existing unlearning methods hard to be applied. To solve this problem, ~\cite{abs-2103-14991} proposes the first graph unlearning framework, GraphEraser. 

\noindent {\bf GraphEraser.} Given an undirected graph $\mathcal{G}_{F}=(\mathcal{V}_{F},\mathcal{E}_{F})$ whose node set $\mathcal{V}_{F}$ consists of a training set $\mathcal{V}$ and a test set $\mathcal{V}_{T}$ (without labels). GraphEraser consists of three phases: (1) balanced graph partition; (2) shard model training; (3) shard model aggregation. Specifically, in step (1), GraphEraser designs two balanced graph partition algorithms (BLPA and BEKM) to get a  partition of the training set $\mathcal{V}$. Different from the vanilla methods such as community detection which are easy to output imbalanced partition, BLPA heuristically assigns the nodes with connections to the same group in a manner similar to Lloyd's algorithm for K-Means clustering until the size of the corresponding group arrives at some threshold. BEKM applies a similar method to the embeddings of graph data to achieve better performance. The balanced partition methods could avoid the case that the imbalanced partition contains large shards whose unlearning process is highly inefficient. Suppose the subgraph held by the $i$-th shard is  $\{ \mathcal{V}_{i} \cup \mathcal{V}_{T}, \mathcal{E}_{i\cup T} \}$, where $\mathcal{E}_{i\cup T}$ is the edge set corresponding to $\mathcal{V}_{i}\cup \mathcal{V}_{T} $. Then in step (2) GraphEraser trains a GNN model for each shard in a transductive manner where the unlabeled test nodes and their incident edges are visible to GNN during training. Then these GNN models are tested on the same graph to predict the labels of transductive test nodes. Considering  these different shard models do not uniformly contribute to the final prediction, in step (3), GraphEraser applies a learning-based aggregation method (LBAggr) to optimize the importance scores of the shard models to improve the global model utility.

\section{Inductive Graph Unlearning}\label{ChallengesinInductive}
\subsection{Problem Definition}
\textbf{Notably, inductive training graphs are different from transductive training graphs.} Given an undirected graph $\mathcal{G}_{F}=(\mathcal{V}_{F},\mathcal{E}_{F})$ whose node set $\mathcal{V}_{F}$ consists of a training set $\mathcal{V}$ and a test set $\mathcal{V}_{T}$, the transductive training graph is $\mathcal{G}_{F}=(\mathcal{V}_{F},\mathcal{E}_{F})$ except the labels of $\mathcal{V}_{T}$, while the inductive training graph is $\mathcal{G}=(\mathcal{V},\mathcal{E})$, where $\mathcal{E}$ is the edge set corresponding to $\mathcal{V}$. Thus, inductive graph unlearning  refers to graph unlearning for inductive training graphs. 

Similar to the transductive setting, we have three types of unlearning requests in the inductive setting: node unlearning, feature unlearning, and edge unlearning. 
\begin{itemize}
    \item For a node unlearning request on node $u \in \mathcal{V}$, the service provider needs to retrain the GNN model on the new training graph $\mathcal{G}_{u}=\mathcal{G} \backslash \{X_{u}, e_{u,v} | \forall v \in \mathcal{N}_{u} \cap \mathcal{V} \}$, where $X_{u}$ represents the attribute of $u$. 
    \item For a feature unlearning request on node $u \in \mathcal{V}$, the service provider needs to retrain the GNN model on the new training graph $\mathcal{G}_{u}=\mathcal{G} \backslash \{X_{u}\}$. 
    \item For an edge unlearning request on edge $e_{u,v} \in \mathcal{E}$, the service provider needs to retrain the GNN model on the new training graph $\mathcal{G}_{u}=\mathcal{G} \backslash \{e_{u,v}\}$.
\end{itemize}

According to the SISA framework, the above three types of unlearning lead to the same unlearning update procedure: update the corresponding training subgraph, retrain the GNN model, and compute the importance scores. This paper mainly focuses on the node unlearning task because it is the most difficult one, where feature unlearning and edge unlearning belong to its subsets.

\begin{figure*}[th]
\centering
    \includegraphics[width=0.9\textwidth,scale=0.5]{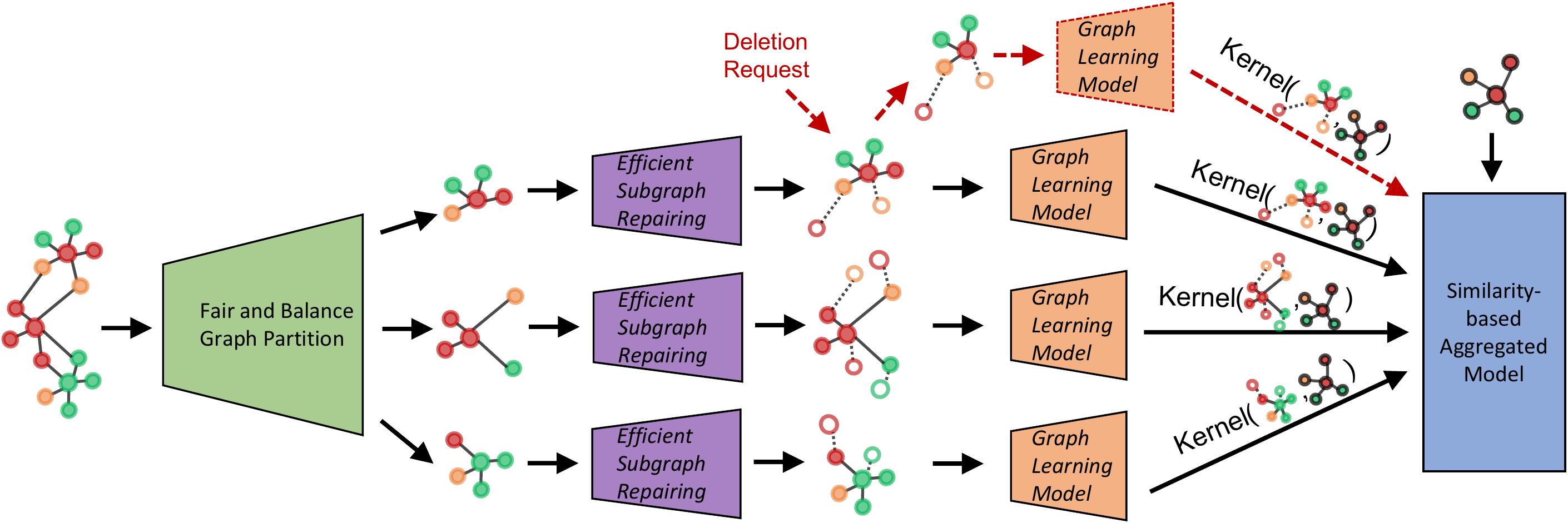}
\caption{ Guided Inductive Graph Unlearning (GUIDE) Framework}
\label{Picture2}
\end{figure*}
\vspace{-0.1in}
\subsection{Challenges}

The main challenges in inductive graph unlearning are caused by the more limited available information in each shard compared with transductive graph unlearning. As we pointed out in Figure~\ref{Picture1}, GraphEraser is unsuitable for the inductive setting. An inductive graph unlearning method should satisfy the following  objectives  simultaneously to achieve satisfactory performance.

\noindent \textbf{C1: Balanced and Fair Graph Partition.} Since there is no help from the test graph during training, we need to make the partition both balanced and fair. A balanced partition makes the retraining time for each shard similar. A fair partition can improve the utility since if several shards are unfair for some classes, their corresponding GNN models would fail to train effective classifiers. Both constraints should be efficiently satisfied for inductive graph learning tasks, where the training graphs may evolve over time or change incrementally. 
    
\noindent \textbf{C2: Comparable Submodel Utility.} Unlike the transductive setting, in our problem we have no information on test nodes, which implies that we lost a lot of information from the original graph data for each subgraph after partitioning the entire graph. Thus, to boost the performance of submodels, we need to restore as much information as possible for each subgraph without using other subgraphs' information (due to the unlearning requirement). 

\noindent \textbf{C3:  Efficient Aggregation Procedure.} Existing learning-based aggregation methods need access to a auxiliary dataset (such as the training graph). Once the  the nodes in the auxiliary data need to be unlearned, the aggregation model must be retrained. Moreover, the importance score of each shard cannot be calculated independently by those aggregation methods, which implies that all importance scores should be updated for optimal values if one shard model is updated, i.e., they are quite inefficient. Therefore, we must design new aggregation methods that assign an importance score for each shard independently and do not rely on additional data. 
    
\section{GUIDE Framework}
\subsection{Overview of GUIDE Framework}

We propose the Guided Inductive Graph Unlearning (GUIDE) framework to achieve the previous objectives. Generally speaking, 
GUIDE consists of three components: guided graph partition with fairness and
balance, efficient subgraph repairing, and similarity-based aggregation. Figure~\ref{Picture2} illustrates the framework of GUIDE. 

\noindent {\bf Guided Graph Partition.} 
To satisfy (C1), we first formulate the problem of finding balanced and fair graph partitions as spectral clustering with linear constraints, which is a quadratic programming problem with binary variables. To solve it efficiently, we relax the constraints and propose the method of GPFB-Fast to solve the relaxed  problem. We then present an improved programming problem via spectral rotation and propose GPFB-SR to solve it. To the best of our knowledge, this is the first study on graph partition under fairness and balance constraints.

\noindent {\bf Efficient Subgraph Repairing.} Such a method aims to satisfy (C2). During the partition process, we retain the original degree information of each node independently (note that this step is independent of future changes of other shards). When the partition is completed, we generate missing neighbors for each node independently according to its features and its original degree information. Specifically, we design three strategies: Zero-Feature Neighbor, Mirror-Feature Neighbor, and MixUp Augmented Neighbor, to reduce the side effects of our graph partition. 

After repairing all subgraphs, the model owner  trains GNN models (in parallel) for all shards isolatedly. The repaired nodes will involve in the GNN message-passing updates. However, the final layer embedding for those repaired nodes will not be used in loss computation.

\noindent {\bf Similarity-based Aggregation.} We develop a similarity-based aggregation method to assign an importance score for each shard independently. The importance score for a shard is calculated by the similarity between its associated subgraph and the test graph. Once a shard is updated, its importance score can be updated efficiently without affecting other shards.

In the following subsections, we will provide details of our three components.

\subsection{Guided Graph Partition with Fairness and Balance}\label{guidepartition}

In this part, we aim to get a partition satisfying the balance and fairness constraints simultaneously. It is notable that such a task is challenging. On the one hand, while some previous work  \cite{abs-2103-14991,ChenWWW22} has proposed some heuristic K-Means clustering variants to achieve balanced graph partitions. Those algorithms are difficult to be extended to graph partitions satisfying two constraints. On the other hand, existing work on fair clustering also does not satisfy the population balance constraint~\cite{Chierichetti0LV17,KleindessnerSAM19,AbbasiBV21}. For further introductions to graph-related clustering, see Appendix~\ref{spectral_scandsr}.

Before showing our method, we first show how to incorporate these two constraints into the graph partition problem. Given a graph dataset $\mathcal{G}=(\mathcal{V},\mathcal{E})$ with all node labels, we suppose $|\mathcal{V}|=n$ and $\mathcal{V} = \dot{\cup}_{s\in [h]} \mathcal{C}_s$ where $\mathcal{C}_s$ denotes the node set with label $s$ (and there are $h$ classes). It is obvious to see that the ratio of label $s$ in the whole dataset is $|\mathcal{C}_s|/n$. Motivated by ~\cite{KleindessnerSAM19},  we can first construct a {\bf label-membership indicator matrix} $\mathbf{F}\in \mathbb{R}^{n \times h}$, where $F_{i,s}=1$ if the label of node $i$ is $s$ and  $F_{i,s} =0$ otherwise. Thus, the sum of entries in the $s$-th column of $\mathbf{F}$ is $|\mathcal{C}_s|$, the number of nodes with label $s$. 
For a given partition $\mathcal{V} = \dot{\cup}_{i\in [v]} \mathcal{V}_i$, we can easily see that the number of nodes with label $s$ in the $i$-th shard is $|\mathcal{C}_s \cap \mathcal{V}_i |$ and its ratio for the $i$-th shard is  $|\mathcal{C}_s \cap \mathcal{V}_i |/|\mathcal{V}_i|$.

In the most balanced case, the sizes of all shards are the same and the size of the $i$-th shard would be $|\mathcal{V}_i|^{*}=n/v$. In the fairest case, the ratio of label $s$ in each shard should be the same as its ratio in the entire dataset, i.e., $(|\mathcal{C}_s \cap \mathcal{V}_i |/|\mathcal{V}_i|)^{*} = |\mathcal{C}_s|/n$. Then when the graph partition satisfies the fairness and balance constraints at the same time, the number of label $s$ in the $i$-th shard should be $|\mathcal{C}_s \cap \mathcal{V}_i |^{*} = \frac{|\mathcal{C}_s|*|\mathcal{V}_i|^{*}}{n} = \frac{|\mathcal{C}_s|}{v}. $\footnote{For simplicity, here we assume $\frac{|\mathcal{C}_s|}{v}$ is an integer. It is easy to extend to general cases.}

The following Theorem illustrates how we transform the fairness constraint and balance constraints to a linear constraint on the group-membership indicator matrix $\mathbf{Y}\in \mathbb{R}^{n\times v}$ (see Section~\ref{spectral_scandsr} in Appendix for the definitions of group-membership indicator matrix $\mathbf{Y}$ and its normalized version $\mathbf{H}$ for a partition). 

\begin{theorem}[Transformation of Fairness and Balance Constraints on Indicator Matrix $\mathbf{Y}$] Based on the previous notations, 
denote the fairness and balance guided matrix by $\mathbf{M}\in \mathbb{R}^{h \times v}$, i.e.,  $M_{s,j} = \frac{|\mathcal{C}_s|}{v}$ denotes the optimal size of label $s$ in the $j$-th shard. For a  partition $\mathcal{V} = \dot{\cup}_{i\in [v]} \mathcal{V}_i$, it is fair and balanced if and only if  
 $\mathbf{F^{\intercal}Y = M}$, where $\mathbf{Y}\in \{0, 1\}^{n\times v}$ is the  group-membership indicator matrix of the partition that has the form in (\ref{e0}).
\label{theorem1}
\end{theorem}

Based on Theorem \ref{theorem1}, it is sufficient for us to find a group-membership
indicator matrix $\mathbf{Y}$ such that $\mathbf{F^{\intercal}Y = M}$. To incorporate into the spectral clustering problem (\ref{e1}), we can leverage  the spectral rotation theory by supposing there is an orthogonal matrix $\mathbf{R} \in \mathbb{R}^{v \times v} $ such that $\mathbf{HR = Y}$ (as illustrated in problem (\ref{e2})). In total, Theorem \ref{theorem1} suggests that to solve the spectral clustering problem with fairness and balance constraints, it is equivalent to solve 

\begin{equation}
\begin{aligned}
\min_{\mathbf{Y\in\mathcal{Y},H,R} } &\quad Tr(\mathbf{H^{\intercal}LH}) \\
    s.t. &\quad \mathbf{H^{\intercal}H = I, F^{\intercal}Y = M, HR = Y,R^TR=I}. 
\end{aligned}
\label{e2_2}
\end{equation}

However, problem (\ref{e2_2}) is a binary quadratic integer programming, which is  hard to solve with low computation cost. By introducing a new balanced and fair guided matrix, we design a new linear constraint on the embedding matrix $\mathbf{H}$ in (\ref{e0}) rather than $\mathbf{Y}$.

\begin{theorem}
[Transformation of Fairness and Balance Constraints on Embedding Matrix $\mathbf{H}$]Denote  the normalized balanced and fair guided matrix by $\mathbf{\widetilde{M}}\in \mathbb{R}^{h \times v}$, i.e., $\mathbf{\widetilde{M}}_{s,j} = \frac{|\mathcal{C}_s|}{\sqrt{nv}}$.

For a  partition $\mathcal{V} = \dot{\cup}_{i\in [v]} \mathcal{V}_i$, it is fair and balanced if and only if $\mathbf{F^{\intercal}H = \widetilde{M}} $, where $\mathbf{H}$ is the normalized group-membership
indicator matrix of the partition which has the form in (\ref{e0}).$\footnote{The omitted proof of Theorem \ref{theorem1} and \ref{theorem2} are provided in Appendix~\ref{appendixproof}.}$

\label{theorem2}
\end{theorem}

Therefore, the optimization problem of finding a graph partition that satisfies the fairness and balance constraints based on RatioCut is 
\begin{equation}
\begin{aligned}
    \min_{\mathbf{H}} &\quad Tr(\mathbf{H^{\intercal}LH}) \quad
    s.t. &\quad \mathbf{H}\in \mathcal{H}, \mathbf{F^{\intercal}H = \widetilde{M}}, 
\end{aligned}
\label{e2_3before}
\end{equation}
where $\mathcal{H}$ is the set of all normalized group-membership indicator matrices. Similar to the standard spectral clustering, we can relax it to 
\begin{equation}
\begin{aligned}
    \min_{\mathbf{H}} &\quad Tr(\mathbf{H^{\intercal}LH}), \quad 
    s.t. &\quad \mathbf{H^{\intercal}H = I, F^{\intercal}H = \widetilde{M}}. 
\end{aligned}
\label{e2_3}
\end{equation}
Problem (\ref{e2_3}) is equivalent to the following problem for a large enough $\alpha$: 
\begin{equation}
\begin{aligned}
    \min_{\mathbf{H}} &\quad Tr(\mathbf{H^{\intercal}LH}) + \alpha \lVert \mathbf{F^{\intercal}H - \widetilde{M}}  \rVert_2^2 \quad s.t. &\quad \mathbf{H^{\intercal}H = I}.
\end{aligned}
\label{e2_4}
\end{equation}
Problem (\ref{e2_4}) can be further written as a quadratic problem over the Stiefel manifold, which can be solved efficiently by the generalized power iteration method~\cite{NieZL17}, i.e., 
\begin{equation}
\begin{aligned}
    \max_{H} &\quad Tr(\mathbf{H}^{\intercal}(\mathbf{W} - \mathbf{D} -\alpha \mathbf{FF}^{\intercal})\mathbf{H} + 2\alpha \mathbf{H^{\intercal}F\widetilde{M}})\\
    s.t. &\quad \mathbf{H^{\intercal}H = I}
\end{aligned}
\label{e2_4_1}.
\end{equation}
After we solve problem (\ref{e2_4_1}) and get the optimal solution $\mathbf{H^{*}}$, we can apply any K-Means clustering  algorithm to its rows to get the final partition of the graph.
The optimization method for problem (\ref{e2_4_1}), Graph Partition with Fairness and Balance (Fast), is summarized into Algorithm \ref{algo1} in Appendix \ref{algorithmspart}.

As pointed out in ~\cite{HuangNH13a}, the obtained relaxed continuous spectral solution could severely deviate from the optimal discrete solution. Motivated by ~\cite{0004NHY17,WangLWNL21}, we add a spectral rotation regularization term to learn better embedding and indicator matrices jointly. In total, we have the following problem. 

\begin{equation}
\begin{aligned}
    \min_{\mathbf{H,Y} } &\quad Tr(\mathbf{H^{\intercal}LH}) + \alpha \lVert \mathbf{F^{\intercal}H - \widetilde{M}}  \rVert_2^2+ \\
    & \quad \beta \lVert \mathbf{HR} - \mathbf{D}^{-\frac{1}{2}}\mathbf{Y(Y^{\intercal}DY)}^{-\frac{1}{2}}  \rVert_2^2 \\
    s.t. &\quad \mathbf{H^{\intercal}H = I, R^{\intercal}R = I}
\end{aligned}.
\label{ae2_4}
\end{equation}
It is notable that as compared with the above problem (\ref{e2_4_1}), we can get an indicator matrix directly without using K-Means clustering algorithms by solving problem (\ref{ae2_4}).  In Appendix ~\ref{solvingalg2}, we show how to solve problem (\ref{ae2_4}) efficiently, and  Algorithm \ref{algo2} in Appendix \ref{algorithmspart} is our final method. When the objective function converges or satisfies certain convergence criteria, we can stop the iteration and get the final indicator matrix $\mathbf{Y}$ satisfying fairness and balance constraints.

\begin{figure*}[h]
\centering
    \includegraphics[width=0.9\textwidth]{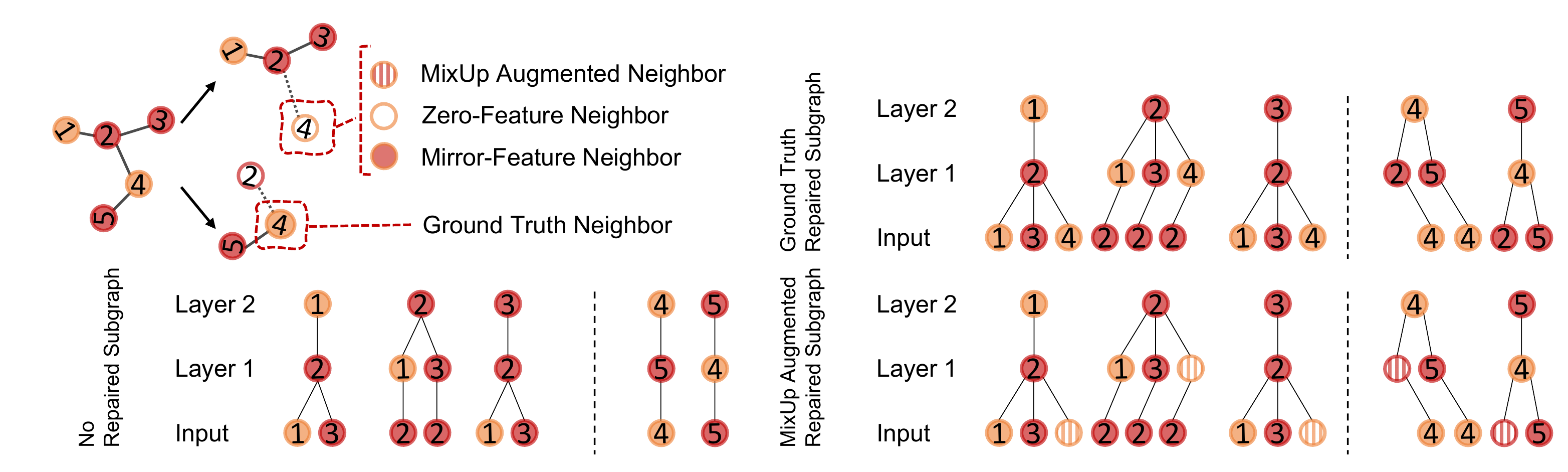}
    \vspace{-0.1in}
\caption{Efficient subgraph repairing. The side effects of partition on the computation graph (message passing operation) of a 2-layer GNN are reduced by the proposed subgraph repairing strategies. The computation graph on a full dataset without partition is provided in Appendix~\ref{SubgraphRepairing}.}
\label{Picture3}
\vspace{-0.1in}
\end{figure*}

\subsection{Efficient Subgraph Repairing}
Subgraph repair has been shown to be helpful in improving the performance of subgraph federated learning~\cite{ZhangYLSY21}. The missing neighbors to be repaired here refer specifically to the 1-hop neighbors of nodes. This is due to the fact that during each training iteration, each node aggregates information from its local (1-hop) neighbors, and as the iterations progress, each node's embedding contains more and more information from further reaches of the graph~\cite{2020Hamilton}. But can these methods really be applied here?

\noindent \textbf{Federated Learning for Missing Neighbors?}  In a subgraph federated learning system, nodes in each subgraph can potentially have connections with those in other subgraphs. To recover these cross-subgraph missing links, ~\cite{ZhangYLSY21} proposes the FedSage+ method to generate the number of missing neighbors and the feature for each missing node. FedSage+ trains a local missing neighbor generation model NeighGen for each local client. The locally computed model gradients of the generative loss are transmitted among the system via the server. Unfortunately, such a federated subgraph repair method involves the training parameters of other clients, which is inapplicable in the setting of graph unlearning.

\noindent
\textbf{Local Generator for Missing Neighbors?} What if we train a neighborhood-generated model on each subgraph? As mentioned by~\cite{ZhangYLSY21}, the federated learning setting is very crucial for the training of NeighGen, which does not hold in graph unlearning. Besides, the additional time cost introduced by NeighGen is very high compared to the training time cost of GNN models. Such complex generative models for subgraph repair cannot be applied in graph unlearning, considering that the primary purpose of graph unlearning is to reduce the retraining time cost.

From the above discussions, we know that an appropriate subgraph repairing method for graph unlearning should satisfy the following properties: (1) It aims to restore the 1-hop neighbors of each node; (2) The repairing procedure should be simple since a complex local generative model will make the unlearning algorithm have high training cost; (3) Due to the unlearning requirement, its repairing procedure for each subgraph cannot rely on other subgraphs' information. 

Motivated by the fact that the insight behind many successful node classification approaches is to explicitly exploit \textit{homophily}, we propose to repair the missing nodes based on their preserved neighbors before partitioning. Generally, homophily refers to the tendency of nodes to share attributes with their neighbors~\cite{annurevsoc415,2020Hamilton}. For example, people tend to form friendships with others who share the same interests. For a preserved node $i$, when homophily exists between its neighbors, we know that its missing neighbors should also have characteristics similar to $\mathbf{x}_i$. When heterogeneity exists, homophily cannot be applied to its neighbors. However, we can still use a simple and effective strategy to repair its local structure.  Specifically, we design the following 
three efficient subgraph repair strategies.

\noindent
\textbf{Zero-Feature Neighbor.} In this approach, each missing neighbor's attribute of node $i$ will be constructed by
$$
\mathbf{\Tilde{x}} = \mathbf{0}_{d\times 1}, 
$$
where $\mathbf{0}_{d\times 1}$ is a $d$-dimentional vector of $0$. As an extreme case where homophily does not exist, we construct $\mathbf{\Tilde{x}}$ without using any information from node $i$'s feature vector, $\mathbf{x}_i$. As we show in Appendix \ref{SubgraphRepairing}, this strategy is sufficient to recover a basic structure of the computation graph. 

\noindent
\textbf{Mirror-Feature Neighbor.} Here each missing neighbor's attribute of node $i$ is constructed by
$$
\mathbf{\Tilde{x}} = \mathbf{x}_i. 
$$
As another extreme case of homophily, we directly copy the feature vector of node $i$ as its missing neighbor's feature. Its repaired computation graph is also shown in Appendix \ref{SubgraphRepairing}.

\noindent
\textbf{MixUp Augmented Neighbor.} 
For a node, the MixUp Augmented Neighbor approach assigns a randomly masked version of the node to its neighbors. In detail, for note $i$, the attribute of each its missing neighbor is constructed as follows.
$$
\mathbf{\Tilde{x}} = \lambda \mathbf{x}_i + (1-\lambda) \mathbf{0}_{d\times 1},
$$
where  $\lambda$ is  randomly sampled from the uniform distribution of $[0,1]$ each time for creating diverse neighbors. MixUp Augmented Neighbor strategy could be considered a trade-off between homophily and heterogeneity. Our strategy seems similar to MixUp~\cite{ZhangCDL18}, which has been used as an efficient data augmentation routine. In short, MixUp extends the training distribution based on the observation that linear interpolations of feature vectors lead to linear interpolations of the associated labels. However, our idea differs from MixUp in that we fix the zero vector in the linear combination and consider only the feature vector, while MixUp requires both features and labels. Directly applying MixUp to repair the missing neighbors of node $i$ requires that node $i$ can provide enough information about the features of its existing neighbors, which is unrealistic after graph partitioning. It is notable that the labels of these newly constructed nodes will not be used during the training process of GNN models. Thus, here we do not need to care what their labels will be.  

The effects of MixUp Augmented Neighbor on the computation graph are shown in Figure \ref{Picture3}. Such simple methods can recover the structure of the computation graph of the GNN model to some extent and with low computation cost. 
\vspace{-0.15in}
\subsection{Similarity-based Aggregation}

Our aggregation method is motivated by recent developments in the interpretability of GNNs. In particular, several GNN explanation studies have been proposed~\cite{YingBYZL19,LuoCXYZC020,SchlichtkrullCT21} and claim that the behavior of GNN models is strongly related to the structure of the training graph. ~\cite{HamiltonYL17} points out that for an inductive GNN model, its generalization to unseen nodes requires "aligning" newly observed subgraphs to the node embeddings on which the algorithm has already been optimized. A new graph with more similar substructures to the training graph is expected to yield better inference results. Thus, we should assign subgraphs which are more similar to the test graph higher importance scores during the inference stage. Here we can directly use graph kernels to measure such similarity. In this paper, we will use the pyramid match graph kernel~\cite{NikolentzosMV17} to compute the similarity score between the test graph and each subgraph, which is a state-of-the-art algorithm for measuring the similarity between unlabeled graphs. \footnote{Note that any similarity measuring algorithm can be used here, depending on the settings of different tasks.} Motivated by our ideas above, we propose our similarity-based aggregation method. 

Specifically, in our method we first represent each graph as a set of vectors corresponding to the embeddings of its vertices in the eigenspace. To find an approximate correspondence between two sets of vectors, we then map these vectors onto multi-resolution histograms and compare these two histograms through a weighted histogram intersection measure~\cite{NikolentzosMV17}.  Given the test graph $G_{t}$ and the subgraph $G_{i}$ of shard $i$ (with depth $L$), denote  $H^{l}_{G_t}$ and $H^{l}_{G_i}$ as the histogram of $G_t$ and $G_i$ at level $l$, respectively. We then calculate the pyramid match kernel over these two histograms: 
\begin{equation}
\begin{aligned}
k(G_t,G_i) = & I(H_{G_t}^{L},H_{G_i}^{L}) + \sum_{l=0}^{L-1} \frac{1}{2^{L-l}} ( I(H_{G_t}^{l},H_{G_i}^{l}) \\
& - I(H_{G_t}^{l+1},H_{G_i}^{l+1})),  
\end{aligned}
\end{equation}
where $I(H_{G_t}^{l},H_{G_i}^{l})$ is the number of nodes that match at level $l$ in the two sets. We refer the readers  to~\cite{NikolentzosMV17} for more details on this kernel. In practice, we can use the \textit{grakel} library ~\cite{JMLRGrakel}\footnote{https://ysig.github.io/GraKeL/} to implement the pyramid match graph kernel.

\subsection{Discussions}
\textbf{Choices of Different Components.} We recommend 
service providers to choose the appropriate partition and subgraph repair methods according to their needs. The choice between GPFB-Fast and GPFB-SR depends on the service provider's preference for graph partitioning: GPFB-SR could lead to a considerably fair and balanced partition, while GPFB-Fast is much faster than GPFB-SR. The choice of subgraph repair strategy depends on the GNN structure we plan to use, as shown in Table~\ref{subgraphrepair}. Zero-Feature Neighbor is more appropriate for the GraphSAGE model, while Mirror-Feature  Neighbor is more appropriate when using the GIN model. The MixUp Augmented method is a general method for all GNNs. 

\noindent
\textbf{Guarantee of Unlearning.} Each component of GUIDE follows the principle of minimizing the use of training graph information. The two proposed graph partitioning algorithms, GPFB-Fast and GPFB-SR, both require only the edge information of nodes with their IDs and labels. The feature information of the nodes is not involved in the graph partitioning step. The subgraph repair procedure uses only the degree information of the entire training graph and the corresponding feature information of each node. The similarity-based aggregation computes the importance score for each shard independently based on the similarity between its corresponding subgraph and the test graph during inference. After receiving an unlearning request, except for the graph partition, both its corresponding shard models and importance scores can be unlearned deterministically. Therefore, similar to SISA \cite{BourtouleCCJTZL21} and GraphEraser \cite{abs-2103-14991}, GUIDE is an approximate unlearning approach. To prove the unlearning ability of GUIDE, we perform the membership inference attack on GUIDE in section~\ref{UnlearnAbility} and  show our results are close to random guessing. These results are consistent with the conclusion of existing work~\cite{BourtouleCCJTZL21,abs-2103-14991,Chen000HZ21}. 

\noindent
\textbf{Computation Complexity Analysis.} For GPFB-Fast, the time cost on initializing $\mathbf{B}$ is $O(nvh)$. In each iteration, the time complexity of updating $\mathbf{P}$ is $O(n^2v+nvh)$, while the time complexity for computing $\mathbf{WH}$ and $\mathbf{FF^{\intercal} H}$ is $O(n^2v)$ and $O(nvh)$ respectively. The complexity of calculating reduced SVD on $\mathbf{P}$ is $O(n^2v)$. The computation cost of K-Means is $O(nv^2)$. Suppose the iteration number of updating $\mathbf{H}$ is $t_1$, then the total computation cost of GPFB-Fast is $O(t_{1}(n^2v+nvh)+nv^2+nvh)$.

For GPFB-SR, the time cost of solving $\mathbf{R}$ is $O(v^3)$, and  the computational complexity for obtaining $\mathbf{Y}$ is $O(nv)$. Therefore, suppose the iteration number of updating $\mathbf{R,H,Y}$ is $t_2$ and the iteration number for obtaining $\mathbf{Y}$ is $t_3$, the total time complexity of GPFB-Rotation is $O(t_{2}(v^3+t_1(n^2v+nvh)+t_3(nv)) +nvh)$.

Although the orders of the time complexity of GPFB-Fast and GPFB-SR both are quadratic in $n$ in theory, the main bottleneck is the matrix computation which can be implemented efficiently in parallel. In Section~\ref{unlearningefficiency}, we will illustrate that in practice the computation costs of GPFB-Fast and GPFB-SR are less than the computation costs of BLPA and BEKM in \cite{abs-2103-14991}, which must be performed sequentially by nodes.

\section{Experimental Results}
We evaluate the performance of GUIDE on the real-world Bitcoin illicit transactions detection task~\cite{BitcoinPaper} and four popular inductive node classification benchmarks~\cite{YangCS16, BojchevskiG18, abs-1811-05868}. 

The evaluation aims to answer the following questions: (1) Unlearning and Implementation Efficiency: How fast can GUIDE handle batch unlearning requests? How efficient are GPFB-Fast and GPFB-SR in practice? (2) Model Utility: Can GUIDE provide state-of-the-art performance for inductive graph learning tasks? (3) Partition Efficacy: Can GPFB-Fast and GPFB-SR output fair and balanced partitions? (4) Efficacy of Subgraph Repairing: Will our subgraph repair strategies help to improve model performance? (5) Efficacy of Similarity-based Aggregation: Can our similarity-based aggregation method reach a level of performance comparable to previous learning-based aggregation methods? (6) Unlearning Ability: Can GUIDE really unlearn the requested nodes? 

\subsection{Experimental Setup}

\noindent
\textbf{Datasets and Experimental Setup.} The Elliptic Bitcoin Dataset~\cite{BitcoinPaper} consists of a time series graph (49 distinct time steps, evenly spaced with an interval of about two weeks) of over 200K bitcoin transactions (nodes) and 234K payment flows (edges) with a total value of \$6 billion. Twenty-one percent of entities (42,019) are labeled licit (exchanges, wallet providers, miners, licit services, etc.). Two percent (4,545) are labeled illicit (scams, malware, terrorist organizations, ransomware, Ponzi schemes, etc.). The remaining transactions are not labeled with regard to licit versus illicit but have other features. A GNN detection model would learn from past transaction graphs and make a prediction for each entity of the new transaction graph. Similar to the temporal split in ~\cite{BitcoinPaper}, which reflects the nature of the task, the first 30 time steps are used to train a GNN model for detecting illicit entities, the next 4  are used for validation, and the last 15 are used for testing. As such, the GNN model is trained in an inductive setting. We set the number of shards for Elliptic to 20, which means that the graph of each time step would be partitioned into 20 subgraphs.

The four popular node classification benchmarks consist of static citation networks and coauthor networks: Cora~\cite{YangCS16}, CiteSeer~\cite{YangCS16}, DBLP~\cite{BojchevskiG18}, and CS~\cite{abs-1811-05868}. The details of four benchmarks are provided in Appendix~\ref{otherdetaildata}. We follow a generally accepted inductive setting in ~\cite{ChenMX18,WuSZFYW19}: we construct one graph containing only training nodes and another graph containing all nodes. Graph partitioning and GNN training are applied to the former one. That means the testing nodes are invisible during the training process. Similar to the setting of ~\cite{abs-2103-14991},  we set the number of shards for Cora, CiteSeer, DBLP, and CS to 20, 20, 100, and 100, respectively, which makes the number of nodes in each shard similar. For all static graph datasets, we randomly split nodes into 80\% and 20\% for training and testing and report the average performance of all models over 10 random splits. In fairness to the evaluation, we also report the  performance of graph unlearning methods on the transductive setting with the same data splitting and model architecture in Appendix~\ref{TransductiveAppendix}.

\noindent
\textbf{Metrics.} For the illicit entity detection task, we opt for two commonly used metrics - AUC and Macro F1 score~\cite{TangLGL22}. AUC measures the area under the ROC Curve. Macro F1 score, the mean of the F1-score of both classes without weighting, provides an objective measure of model performance in the face of extreme class imbalance. For inductive node classification benchmarks, we consider classification accuracy as in ~\cite{ChenMX18,WuSZFYW19}. 

To measure the quality of a graph partition, we design two partition metrics: balance score and fairness score. In the following, we provide the
definitions of balance score and fairness score for a partition $\{\mathcal{V}_i\}_{i=1}^v$ of the graph $(\mathcal{V}, \mathcal{E})$ with number of nodes $n$. 

\noindent
\textbf{Balance Score:} Denote the optimal size of the $i$-th shard as  $|\mathcal{V}_i|^{*} = \frac{n}{v}$. To quantify the degree of balance,  we formally define its population balance score as follows: 
$$
\mathcal{B}_b = -\frac{1}{2} \Sigma_{i=1}^{v} \frac{||\mathcal{V}_i| - |\mathcal{V}_i|^{*}|}{n},
$$ 
where $ -1 \le \mathcal{B}_b \le 0$. In the optimal case, $||\mathcal{V}_i| - |\mathcal{V}_i|^{*}|=0$ for all $i\in [v]$, which implies that $\mathcal{B}_b=0$. When the partition of the $i$-th shard is unbalanced, we have $||\mathcal{V}_i| - |\mathcal{V}_i|^{*}| > 0$, i.e., $\mathcal{B}_b<0$. We can also easily see that larger  $\mathcal{B}_b$ indicates that the partition is  more balanced.

\noindent
\textbf{Fairness Score:} Denote the node set with label $s$ as $\mathcal{C}_s$ for $s\in [h]$, we have $\mathcal{V} = \dot{\cup}_{s\in [h]} \mathcal{C}_s$. It is easy to know that the ratio of nodes with label $s$ in the full dataset is $\frac{|\mathcal{C}_s|}{|\mathcal{V}|}$. Similarly, the ratio of nodes with label $s$ in the $i$-th shard is  $\frac{|C_s\cap \mathcal{V}_i |}{|\mathcal{V}_i |}$. The fairness score can be computed by 
$$
\mathcal{B}_f = -\frac{1}{2v} \sum_{i=1}^{v} \sum_{s=1}^{h} \bigl\vert \frac{|C_s \cap V_i |}{|V_i|} - \frac{|C_s|}{n} \bigr\vert,
$$
where $ -1 \le \mathcal{B}_f \le 0$. In the fairest case, the ratios for every class over all shards are equal, i.e., $\frac{|C_s\cap \mathcal{V}_i |}{|\mathcal{V}_i |} = \frac{|\mathcal{C}_s|}{|\mathcal{V}|}$ for all $ i\in [v]$, which implies that $\mathcal{B}_f=0$. When the class $s$ in the $i$-th shard is unfair, we have $|\frac{|C_s\cap \mathcal{V}_i |}{|\mathcal{V}_i |} - \frac{|\mathcal{C}_s|}{|\mathcal{V}|}| > 0 $ so that $\mathcal{B}_f<0$. Moreover, we can see a larger $B_f$ indicates the partition is fairer.

\noindent
\textbf{Baselines.} We compare GUIDE with two standard baselines (Scratch, Random) and two graph unlearning methods (Eraser-BLPA, Eraser-BEKM). For the fraud detection task, we apply graph unlearning methods on a designed illicit entity detection GNN model. For inductive node classification task, we apply graph unlearning methods on 6 popular inductive GNN models to compare their efficiency and model utility, including GraphSAGE~\cite{HamiltonYL17}, GIN~\cite{XuHLJ19}, GAT~\cite{VelickovicCCRLB18}, GATv2~\cite{Brody0Y22}, SuperGAT~\cite{KimO21}, APPNP~\cite{KlicperaBG19}. The detailed settings of those baselines and GNN models are reported in Appendix~\ref{otherdetailbaseline}.

For the implementation of GUIDE, we first apply GPFB-Fast or GPFB-SR on the training graph. The partitioned subgraphs are then repaired by our proposed graph repair strategies. After training the GNN model for each shard independently, we compute an importance score for each shard using the similarity-based aggregation. We name the two implementations of GUIDE (with different partition methods) as  GUIDE-Fast (GUIDE with GPFB-Fast) and GUIDE-SR (GUIDE with GPFB-SR) for convenience, respectively.

For both GPFB-Fast and GPFB-SR, the regularization parameter $\alpha$ is determined via grid search from $\{0.0001,0.001,0.01\}$. For GPFB-SR, the regularization parameter $\beta$ is determined by grid search from $\{1, 2, 3, 4, 5\}$. Unless otherwise indicated, we take the MixUp Augmented Neighbor as our default repairing strategy. The performances of Zero-Feature Neighbor and Mirror-Feature Neighbor are also reported. 
We use the pyramid match graph kernel to compute the similarity score between each repaired subgraph and the test graph. But we argue that any method of measuring the similarity between graphs can be applied here. 

\noindent
\textbf{Implementation.} All experiments are conducted on a server with 128G memory, two NVIDIA RTX 3090 GPUs with 24GB RAM, and Ubuntu 20.04 LTS OS.

\subsection{Unlearning and Implementation Efficiency}\label{unlearningefficiency}

\begin{figure}[th]
\centering
\begin{subfigure}{0.45\textwidth}
 \centering
 \includegraphics[width=\textwidth]{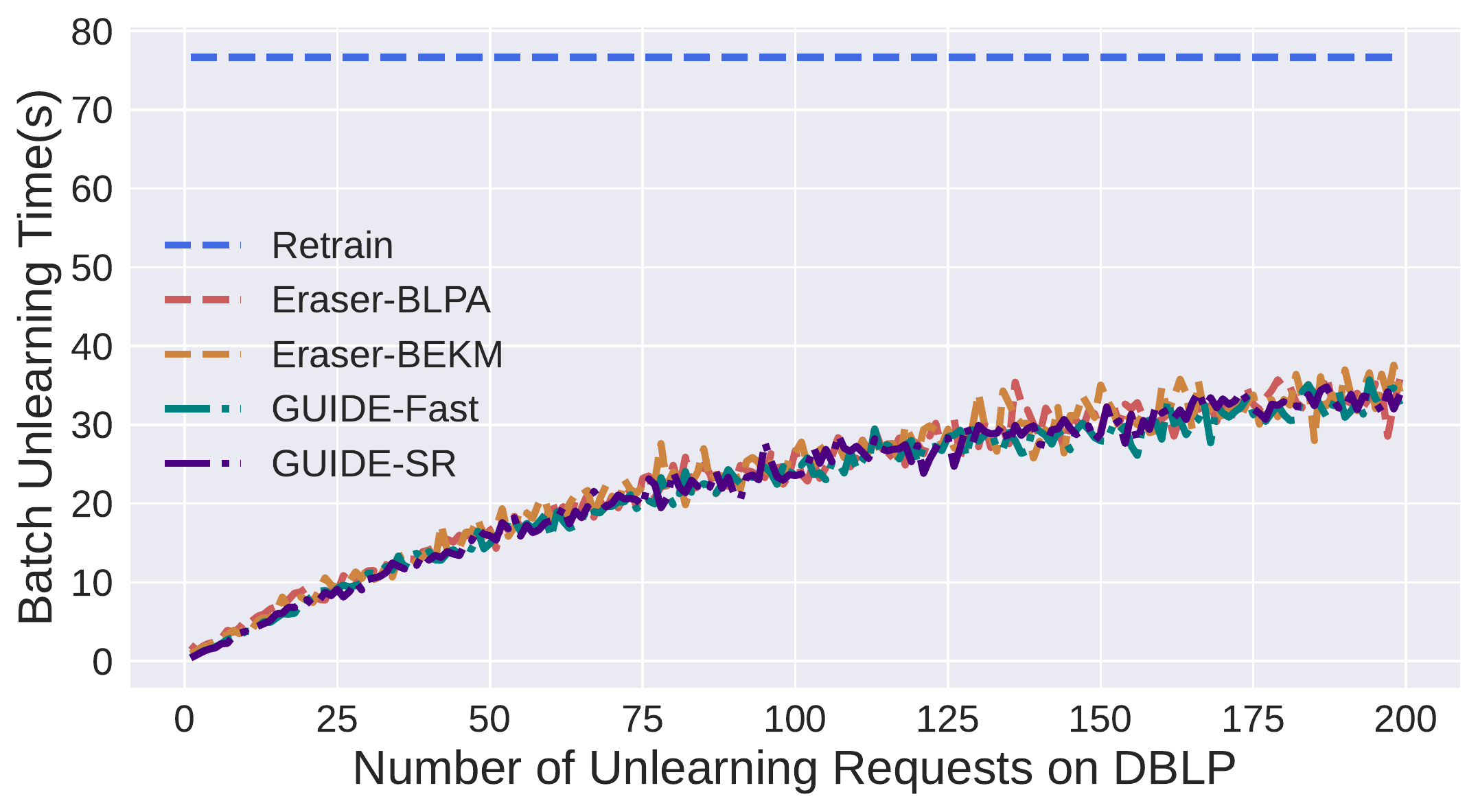}
\end{subfigure}
\hfill
\begin{subfigure}{0.45\textwidth}
 \centering
  \includegraphics[width=\textwidth]{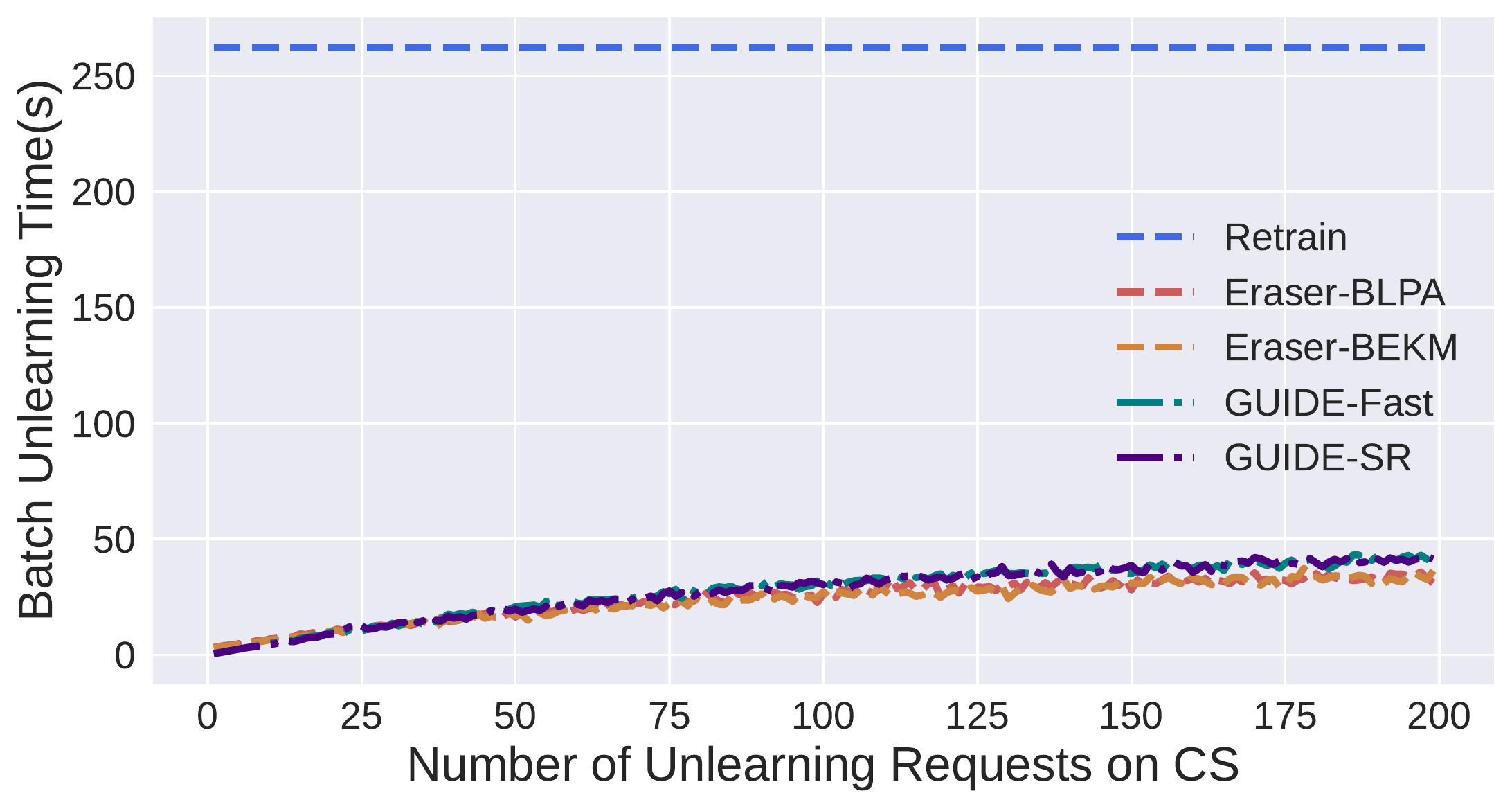}
\end{subfigure}
\hfill
\begin{subfigure}{0.45\textwidth}
 \centering
  \includegraphics[width=\textwidth]{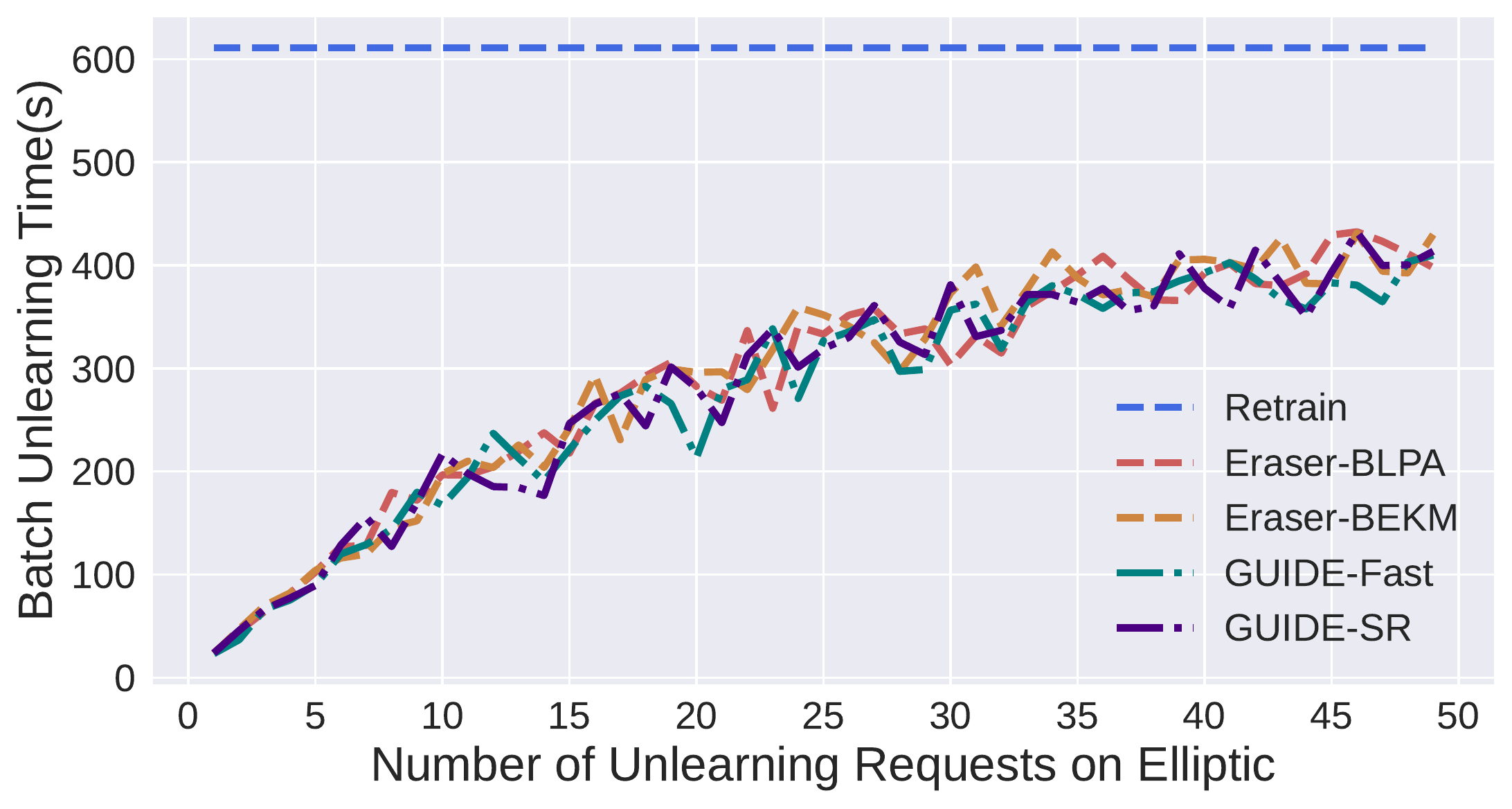}
\end{subfigure}
\caption{Batch unlearning time on large-scale graph datasets.}
\label{PictureBatchUnlearning}
\vspace{-0.1in}
\end{figure}

\textbf{Batch Unlearning Time.} We compare the batch unlearning time of GUIDE and GraphEraser on three graph datasets. The time of Scratch is also reported as the baseline. Our results are shown in Figure~\ref{PictureBatchUnlearning}. We can see that as the number of unlearning nodes increases, more and more shards are involved, so it will take a longer time to unlearn. When all shards need to be updated, the unlearning time tends to be stable. However, since the size of each subgraph is small, it is still faster than retraining from scratch on a large graph. The interesting point is that GUIDE is expected to have a lower unlearning time than GraphEraser due to a more balanced partition and independent importance score updates, but as shown in Appendix~\ref{AveBatchUnlearning}, we can only observe such a trend when the batch size of unlearning is small.  The reason is that subgraph repair makes each subgraph's size larger than its original size. The actual training time of each submodel may be higher than its training time on a smaller subgraph without repairing. The submodel training time will dominate the unlearning time when the unlearning batch size is large. However, we claim that such a trade-off between unlearning efficiency and model utility is reasonable because the unlearning efficiency degrades slightly in the comparison, while the model utility gets a significant improvement (as we will show later).

\noindent
\textbf{Implementation Time.} In the inductive setting, the GNN model should learn continuously or keep life-long learning based on those incremental samples. Therefore, implementation efficiency is especially important when facing evolving graphs or multi-graphs. In the following, we report the graph partition time cost for four methods in Table~\ref{Partitiontime}. 

It is notable that the results in~\cite{abs-2103-14991} follow a different setting compared to our experiments. \cite{abs-2103-14991} sets the number of shards on CS to $k=30$ and uses a pre-trained GNN model to generate node embeddings for BEKM in the transductive setting. In our setting, the number of shards on CS is $k=100$. Following the requirements of the inductive setting, we generate  node embeddings with the default setting of BEKM, which is time-consuming when the dataset size is large.
For the Elliptic dataset, we partition its temporal transaction graphs separately according to their timestamps. As observed in Table~ \ref{Partitiontime}, GPFB-Fast takes the shortest time for partition. As explained in section~\ref{guidepartition}, GPFB-Fast is simple to implement and can be solved efficiently by  using standard linear algebra software. For GPFB-SR, we can see it is always faster than BEKM and is comparable with BLPA in some cases. Moreover, it is  slower than GPFB-Fast, which is reasonable  as it needs more iterations to find a better solution. 

\begin{table}[t]
\begin{threeparttable}
\caption{Graph partition time of 4 methods(s).}
\begin{tabular*}{\linewidth}{@{\extracolsep{\fill}}lrr;{1pt/1pt}rr} 
\hline\hline
\multicolumn{1}{l}{Dataset} & \multicolumn{1}{r}{BLPA} & \multicolumn{1}{r;{1pt/1pt}}{BEKM} & \multicolumn{1}{r}{GPFB-Fast}     & \multicolumn{1}{r}{GPFB-SR}               \\ 
\hline\hline
\multicolumn{1}{l}{Cora}              & 5.41    & 10.10        & \bf{0.24}  	 & 2.85    \\
\multicolumn{1}{l}{CiteSeer}          & 6.36    & 14.56        & \bf{0.31} 	 & 3.54  \\ 
\multicolumn{1}{l}{CS}                & 38.77   & 5454.36      & \bf{15.71}   & 40.02    \\
\multicolumn{1}{l}{DBLP}              & 37.30   & 5182.10      & \bf{14.44}   & 33.52    \\
\multicolumn{1}{l}{Elliptic}          & 303.02  & 1089.72     & \bf{26.19}   & 201.99    \\
\hline\hline
\end{tabular*}
\label{Partitiontime}
\begin{tablenotes}
   \item[*] The huge increase in BEKM's computation cost comes from its linear relationship with the number of shards and node embedding generation process. 
  \end{tablenotes}
\end{threeparttable}
\end{table}

\begin{figure}[t]
\centering
\begin{subfigure}{0.48\textwidth}
 \centering
 \includegraphics[width=\textwidth]{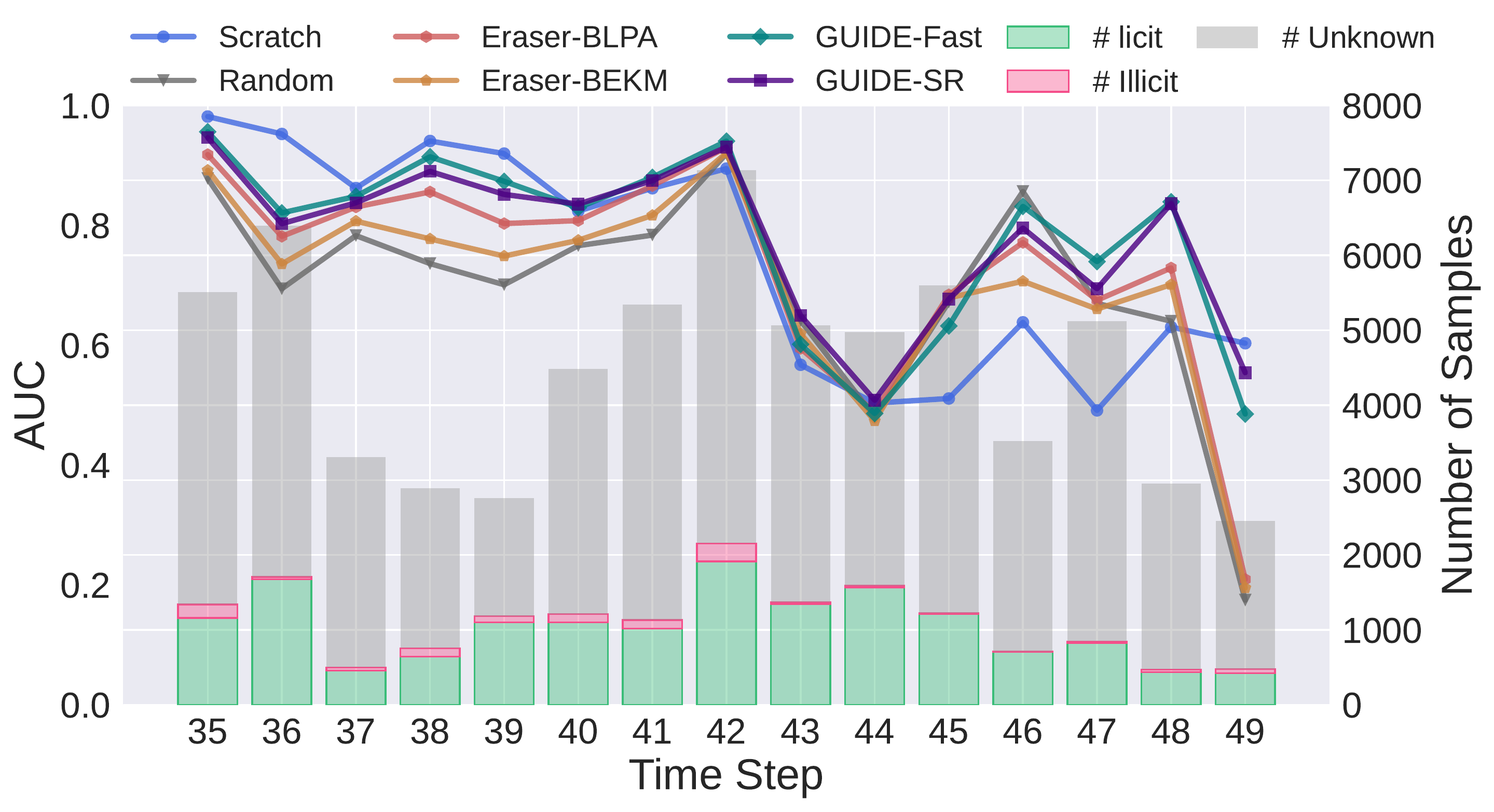}
\end{subfigure}
\hfill
\begin{subfigure}{0.48\textwidth}
 \centering
  \includegraphics[width=\textwidth]{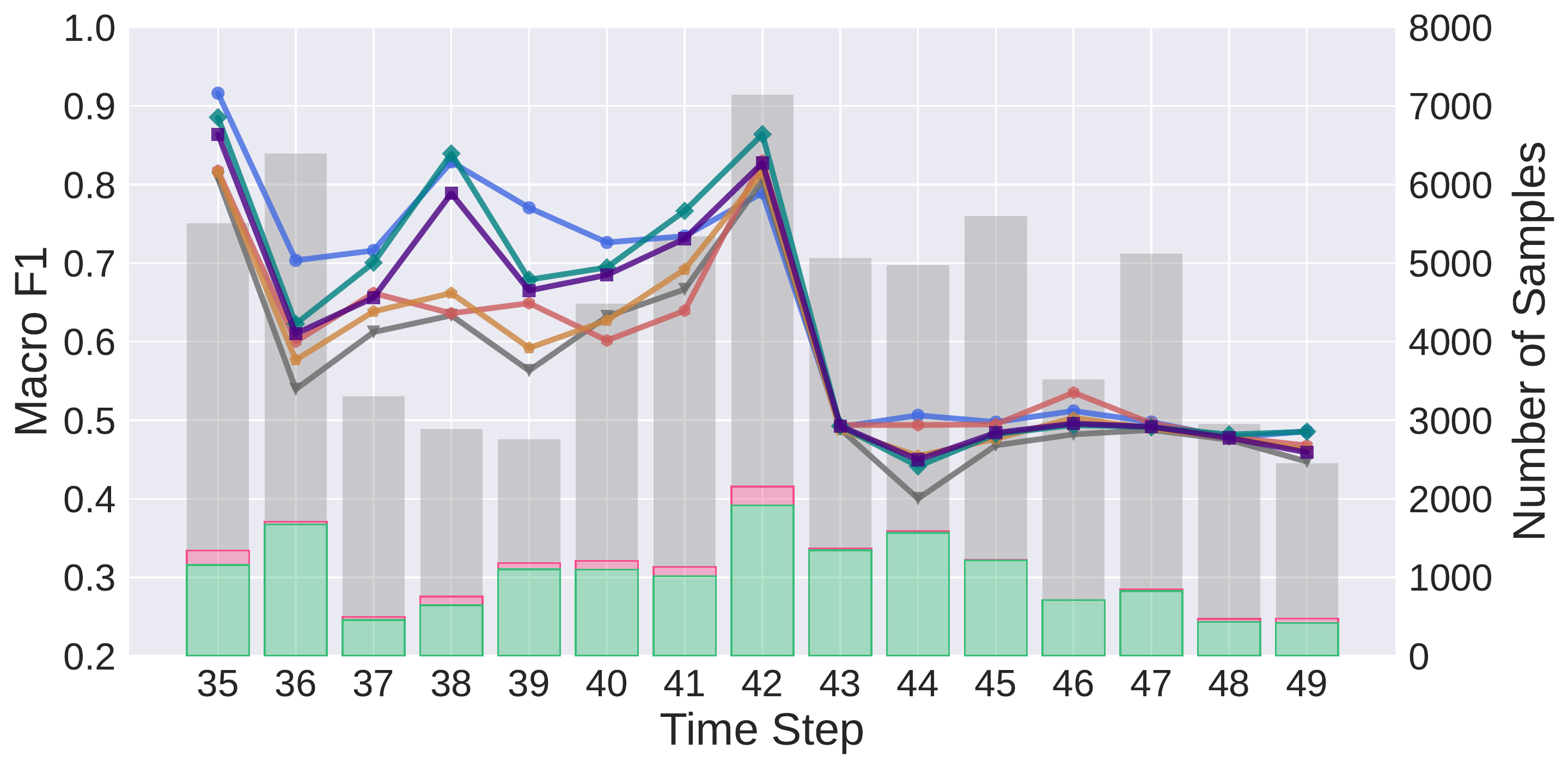}
\end{subfigure}
\caption{Illicit entity detection results over test time span.}
\label{Picturebitcoin}
\vspace{-0.1in}
\end{figure}

\begin{table*}[ht]
\begin{threeparttable}
\centering
\caption{Node classification accuracy of 5 graph unlearning methods with 6 inductive GNN models (\%).}
\vspace{-0.1in}
\begin{tabular*}{\linewidth}{@{\extracolsep{\fill}}ll;{1pt/1pt}c;{1pt/1pt}ccc;{1pt/1pt}cc} 
\hline\hline
\multicolumn{1}{l}{Dataset}                   & Model    & Scratch                             & Random                                       & Eraser-BLPA                                         & Eraser-BEKM                                         & GUIDE-Fast                                    & GUIDE-SR                                 \\ 
\hline\hline
\multicolumn{1}{l}{\multirow{6}{*}{Cora}} & SuperGAT  & 89.17$\pm$0.00 & 31.57$\pm$0.04          & 41.74$\pm$0.16          & 44.92$\pm$0.57          & 65.69$\pm$0.04          & \bf{66.49$\pm$0.12}  \\
                          & GATv2    & 88.94$\pm$0.00 & 31.22$\pm$0.04          & 43.83$\pm$0.68          & 36.62$\pm$0.55          & 66.80$\pm$0.08          & \bf{68.10$\pm$0.16}  \\
                          & SAGE     & 92.73$\pm$0.00 & 53.68$\pm$0.18          & 44.20$\pm$0.37          & 53.57$\pm$0.60          & 71.33$\pm$0.10          & \bf{72.26$\pm$0.04}  \\
                          & GIN      & 87.07$\pm$0.13 & 56.49$\pm$0.26          & 67.84$\pm$0.14          & 65.55$\pm$0.29          & 76.40$\pm$0.05          & \bf{77.06$\pm$0.06}  \\
                          & GAT      & 88.97$\pm$0.00 & 31.90$\pm$0.07          & 38.91$\pm$0.36          & 34.10$\pm$0.34          & 66.25$\pm$0.09          & \bf{66.40$\pm$0.09}  \\
                          & APPNP     & 85.96$\pm$0.03  &51.28$\pm$0.13   &38.02$\pm$0.26   &46.38$\pm$0.12   &64.14$\pm$0.07   &\bf{64.56$\pm$0.05}  \\
\hdashline[1pt/1pt]
\multicolumn{1}{l}{\multirow{6}{*}{CiteSeer}} & SuperGAT & 79.33$\pm$0.00 & 25.44$\pm$1.34          & 53.31$\pm$1.15          & 45.98$\pm$0.48          & 70.66$\pm$0.02          & \bf{71.17$\pm$0.02}  \\
                          & GATv2    & 79.53$\pm$0.00 & 25.88$\pm$1.45          & 58.50$\pm$0.36          & 41.04$\pm$1.58          & 70.78$\pm$0.02          & \bf{71.26$\pm$0.02}  \\
                          & SAGE     & 83.08$\pm$0.00 & 69.10$\pm$0.05          & 66.90$\pm$0.06          & 69.25$\pm$0.05          & \bf{72.71$\pm$0.02} & 72.38$\pm$0.01           \\
                          & GIN      & 81.20$\pm$0.06 & 58.02$\pm$0.41          & 66.29$\pm$0.11          & 64.21$\pm$0.13          & 69.64$\pm$0.07          & \bf{69.67$\pm$0.04}  \\
                          & GAT      & 79.61$\pm$0.00 & 26.32$\pm$1.46          & 58.57$\pm$0.64          & 43.46$\pm$1.17          & 70.66$\pm$0.02          & \bf{71.02$\pm$0.02}  \\
                          & APPNP     & 77.49$\pm$0.00  &72.98$\pm$0.02   &66.33$\pm$0.40   &71.29$\pm$0.04   &73.09$\pm$0.03   &73.43$\pm$0.02  \\
\hdashline[1pt/1pt]
\multicolumn{1}{l}{\multirow{6}{*}{DBLP}} & SuperGAT & 84.21$\pm$0.00 & 44.67$\pm$0.00          & 70.27$\pm$0.01          & 69.84$\pm$0.01          & \bf{71.67$\pm$0.01} & 69.29$\pm$0.01           \\
                          & GATv2    & 83.93$\pm$0.00 & 44.67$\pm$0.00          & 70.23$\pm$0.01          & 69.06$\pm$0.05          & \bf{71.69$\pm$0.01} & 69.10$\pm$0.00           \\
                          & SAGE     & 86.72$\pm$0.00 & 60.38$\pm$0.02          & 70.13$\pm$0.00          & 69.70$\pm$0.00          & 71.92$\pm$0.01          & \bf{72.16$\pm$0.01}  \\
                          & GIN      & 87.35$\pm$0.01 & 67.76$\pm$0.02          & \bf{79.09$\pm$0.02} & 75.78$\pm$0.09          & 77.11$\pm$0.03          & 77.51$\pm$0.00           \\
                          & GAT      & 84.05$\pm$0.00 & 44.67$\pm$0.00          & 70.41$\pm$0.01          & 68.51$\pm$0.08          & \bf{71.39$\pm$0.01} & 68.70$\pm$0.01           \\
                          & APPNP     & 83.80$\pm$0.00  &67.53$\pm$0.00   &71.56$\pm$0.01   &70.96$\pm$0.01   &\bf{73.62}$\pm$0.01   &72.84$\pm$0.01  \\
\hdashline[1pt/1pt]
\multicolumn{1}{l}{\multirow{6}{*}{CS}}       & SuperGAT & 87.57$\pm$0.00 & 22.79$\pm$0.01          & 53.01$\pm$0.02          & 41.98$\pm$0.25                                            & \bf{69.63$\pm$0.00} & 69.53$\pm$0.01           \\
                          & GATv2    & 86.98$\pm$0.00 & 22.79$\pm$0.01          & 53.58$\pm$0.04          & 40.08$\pm$0.29          & \bf{73.28$\pm$0.01} & 73.15$\pm$0.01           \\
                          & SAGE     & 91.79$\pm$0.00 & 71.96$\pm$0.02          & 57.37$\pm$0.04          & 74.38$\pm$0.01          & \bf{80.68$\pm$0.00} & 80.67$\pm$0.00           \\
                          & GIN      & 83.69$\pm$0.18 & 36.70$\pm$0.01          & 75.42$\pm$0.15          & \bf{83.65$\pm$0.01} & 79.24$\pm$0.01          & 79.73$\pm$0.02           \\
                          & GAT      & 87.37$\pm$0.00 & 22.79$\pm$0.01          & 53.24$\pm$0.01          & 43.17$\pm$1.04          & \bf{69.55$\pm$0.01} & 69.45$\pm$0.01           \\
                          & APPNP     & 78.70$\pm$0.01  &58.03$\pm$0.01   &48.24$\pm$0.10   &47.81$\pm$0.09   &74.38$\pm$0.01   &\bf{74.44$\pm$0.01}  \\
\hdashline[1pt/1pt]
\multicolumn{2}{l;{1pt/1pt}}{Normalized Score} & 100.00 & 0.00 &20.42	&23.71	&59.52	&59.40\\
\hline\hline
\end{tabular*}
\label{modelutility}

\end{threeparttable}
\end{table*}

\subsection{Model Utility}

\subsubsection{Fraud Detection}

We construct a GNN model with three GINconv layers to conduct the illicit entity detection task. Each GINconv layer consists of a 3-layer MLP. After tuning the hyperparameters based on the validation data, we set the size of node embedding to 1024. The model is trained with 200 epochs. The performance of different graph unlearning methods based on our GNN model is shown in Figure~\ref{Picturebitcoin}. It is easy to see that GUIDE performs very close to Scratch for two metrics during the first 8 time steps, and there is a clear performance gap between GUIDE and others. Especially in the 38th time step, GUIDE outperforms other methods with more than 10\% Macro F1 score. During the last 7 time steps, all methods provide similar Macro F1 scores due to the very limited illicit samples in the test graph, while GUIDE still produces higher AUC scores than other methods.

\subsubsection{Inductive Node Classification}

We evaluate the model utility of different graph unlearning methods on 6 commonly used inductive GNN models. Table~\ref{modelutility} presents the average results for these methods on four graph datasets. Comparing the results of Scratch and Random, we can find that there is a large gap between them. Most of the time, the node classification accuracy of the Random method is less than half of the Scratch method. Taking this gap as $100\%$, we can calculate normalized scores of the results given by other graph unlearning methods to quantify the improvement of those methods to the Random method. We can see the improvement of GraphEraser methods to the Random method is only $20\%$. It is not surprising because the available information is very limited in the inductive setting. Thus, we can see GraphEraser is unsuitable for the inductive setting. However,  we also find that the GIN model can achieve the highest node classification accuracy with the help of GraphEraser sometimes (e.g., over the CS data). This is mainly due to the unbalanced partition since the learning-based aggregation assigns a small score to the shards with a small size. But we argue that it is not advisable to sacrifice too much balance for better performance since it will increase the unlearning time cost. Compared to the results of GraphEraser, GUIDE achieves the best performance for almost all models on four datasets. The normalized scores of GUIDE-Fast and GUIDE-SR are both $\sim 2\times$ higher than the results of GraphEraser.

In comparison to the Certified Graph Unlearning method~\cite{abs-2206-09140}, we apply the SGC model to the Scratch method and five graph unlearning methods. The results are provided in Appendix~\ref{SGDAppendix}, showing that there is a large gap between the performance of SGC and the performance of state-of-the-art GNNs for inductive graph learning tasks. We also report the results of a 2-layer MLP (Multi-Layer Perceptron) model without considering the graph structure in Appendix~\ref{SGDAppendix}.

\vspace{-0.1in}
\subsection{Partition Efficacy}

\begin{figure*}[ht]
\centering
    \includegraphics[width=\textwidth]{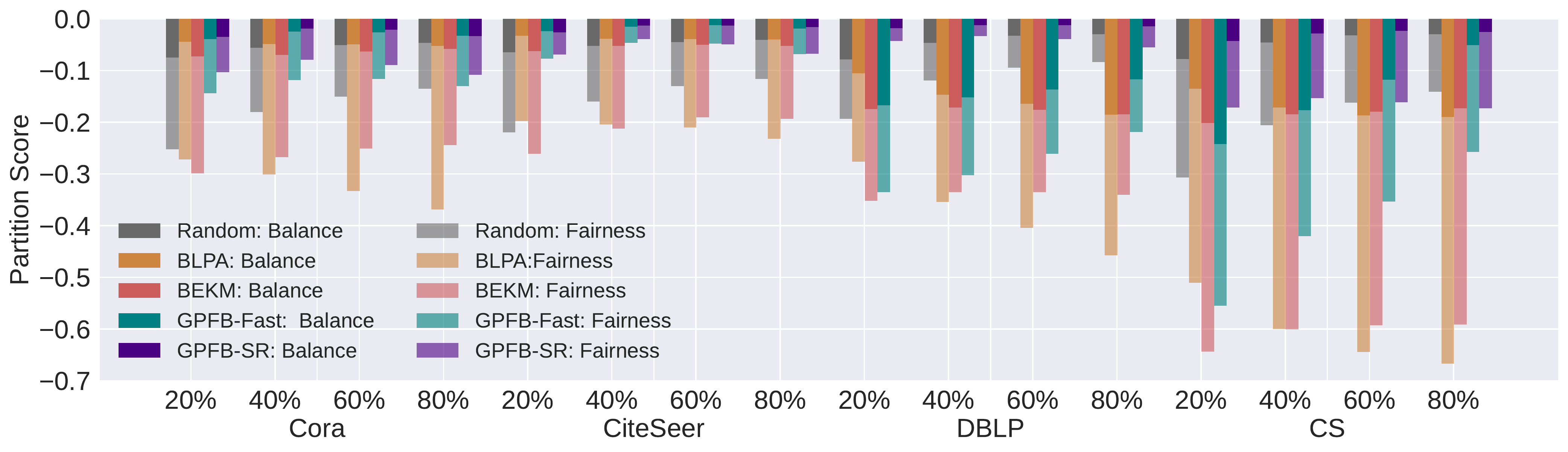}
    \vspace{-0.3in}
\caption{Partition scores of 5 methods on different datasets.}
\label{Picture4}
\vspace{-0.1in}
\end{figure*}

We can quantify the partition efficacy of  graph unlearning methods by calculating the balance score and fairness score of each partition. The average results on different parts (20$\%$, 40$\%$, 60$\%$, 80$\%$) of four datasets are presented in Figure~\ref{Picture4}, where the partition score is the summation of the balance score and the fairness score. A smaller absolute value of this negative score indicates that the corresponding partition 
is fairer and more balanced. As we can see from Figure~\ref{Picture4}, the performance of GPFB-SR is always comparable with the performance of Random. The partition scores of GPFB-SR are $\sim 3\times$ better than the scores of BLPA and BEKM. We also present the distribution of shard sizes in Appendix~\ref{PartitionSize}, which supports this claim. Although the partition scores of GPFB-Fast are worse than those of GPFB-SR, they are almost always better than the results of BLPA and BEKM. The results demonstrate that GUIDE could bring about partitions with balance and fairness,  achieving satisfactory performance.  

\begin{table*}[th]
\centering
\caption{Results of different subgraph repairing strategies on Cora(\%)}
\begin{tabular*}{\textwidth}{@{\extracolsep{\fill}}c;{1pt/1pt}l;{1pt/1pt}l;{1pt/1pt}l;{1pt/1pt}lll} 
\hline\hline
\multicolumn{1}{l;{1pt/1pt}}{Partition Method}               & Model & \multicolumn{1}{c;{1pt/1pt}}{Ground Truth} & \multicolumn{1}{c;{1pt/1pt}}{No Repairing} & \multicolumn{1}{c}{Mirror Feature} & \multicolumn{1}{c}{Zero Feature} & \multicolumn{1}{c}{MixUp}  \\ 
\hline\hline
\multirow{3}{*}{GPFB-Fast}     & SAGE  & 77.26$\pm$0.04                             & 59.98$\pm$0.19                             & 63.22$\pm$0.08                     & 73.55$\pm$0.05                   & 71.33$\pm$0.10             \\
                               & GIN   & 79.26$\pm$0.04                             & 70.09$\pm$0.05                             & 77.13$\pm$0.07                     & 72.07$\pm$0.06                   & 76.40$\pm$0.05             \\
                               & GAT   & 70.52$\pm$0.10                             & 49.63$\pm$0.08                             & 62.90$\pm$0.09                     & 66.52$\pm$0.07                   & 66.25$\pm$0.09             \\ 
\hdashline[1pt/1pt]
\multirow{3}{*}{GPFB-SR} & SAGE  & 77.78$\pm$0.02                             & 59.98$\pm$0.09                             & 65.67$\pm$0.08                     & 74.38$\pm$0.04                   & 72.26$\pm$0.04             \\
                               & GIN   & 78.96$\pm$0.02                             & 69.28$\pm$0.14                             & 75.16$\pm$0.09                     & 72.20$\pm$0.06                   & 77.06$\pm$0.06             \\
                               & GAT   & 70.85$\pm$0.06                             & 50.00$\pm$0.20                             & 65.18$\pm$0.09                     & 67.08$\pm$0.08                   & 66.40$\pm$0.09             \\ 
\hdashline[1pt/1pt]
\multicolumn{2}{l;{1pt/1pt}}{Normalized Score}    & \multicolumn{1}{c;{1pt/1pt}}{100.00}       & \multicolumn{1}{c;{1pt/1pt}}{0.00}         & \multicolumn{1}{c}{54.09}          & \multicolumn{1}{c}{62.32}        & \multicolumn{1}{c}{73.65}  \\
\hline\hline
\end{tabular*}
\label{subgraphrepair}
\end{table*}

\vspace{-0.1in}
\subsection{Efficacy of Subgraph Repairing}
To test the efficacy of our proposed subgraph repair strategies, we compare the performance of our three strategies with the ground truth subgraphs and the subgraphs without repairing on the Cora dataset as an ablation study. As shown in Table~\ref{subgraphrepair}, all three subgraph repairing strategies are helpful in improving model performance. The simplest Zero-Feature Neighbor could achieve a $62.32\%$ improvement. It is not surprising that Mirror-Feature Neighbor behaves worse than Zero-Feature Neighbor since the contributions of the aggregated information from the Mirror-Feature Neighbor are zero. Here we randomly select $\lambda \in [0,1]$ for each missing node to generate the mix-up between the zero and mirror feature. Considering the real application where heterogeneous neighbors may not share the same feature, we can also control this mix-up process by randomly selecting $\lambda \in [0,\tau]$, where $\tau \in [0,1)$ can be decided by testing on a small subset.

\vspace{-0.1in}
\subsection{Efficacy of Similarity-based Aggregation}

\begin{table}[t]
\centering
\caption{Results of 3 aggregation methods on Cora(\%).}
\begin{tabular*}{\linewidth}{@{\extracolsep{\fill}}ll;{1pt/1pt}lll} 
\hline\hline
\multicolumn{2}{c;{1pt/1pt}}{Method}                                            & \multicolumn{1}{c}{Average}         & \multicolumn{1}{c}{LBAggr}          & \multicolumn{1}{c}{SimiAgg}         \\ 
\hline\hline
\multirow{3}{*}{\begin{tabular}[c]{@{}l@{}}Fast\end{tabular}}     & SAGE & 71.11$\pm$0.10 & 69.02$\pm$0.17 & \bf{71.33$\pm$0.10}  \\
                                                                         & GIN  & \bf{76.40$\pm$0.06} & 75.25$\pm$0.08 & \bf{76.40$\pm$0.05}  \\
                                                                         & GAT  & 66.10$\pm$0.09 & 65.97$\pm$0.18 & \bf{66.25$\pm$0.09}  \\ 
\hdashline[1pt/1pt]
\multirow{3}{*}{\begin{tabular}[c]{@{}l@{}}SR\end{tabular}} & SAGE & 72.07$\pm$0.05 & 70.55$\pm$0.16 & \bf{72.26$\pm$0.04}  \\
                                                                         & GIN  & 76.69$\pm$0.06 & 76.32$\pm$0.03 & \bf{77.06$\pm$0.06}  \\
                                                                         & GAT  & 66.01$\pm$0.09 & \bf{67.15$\pm$0.20} & 66.40$\pm$0.09  \\
\hline\hline
\end{tabular*}
\label{Aggregation}
\end{table}
To illustrate the performance of the similarity-based aggregation, we compare it with the average aggregation and the learning-based aggregation methods. We aggregate the predictions of GNN models trained based on the partition of GPFB-Fast and GPFB-SR. For convenience, we denote the two partition methods as 'Fast' and 'SR' respectively in Table~\ref{Aggregation}. Even though we train the LBAggr on the full training graph, its performance is not as stunning as SimiAgg. The reason may be caused by the inductive setting, where the behaviors of those submodels on the training subgraphs may differ from those on the test graph. But still the differences between the three aggregation methods are quite small. It is because the fair and balance graph partition and subgrpah repair have improved each subgraph, leading to an improved submodel in each shard. Thus SimAgg isn't significantly better than the average weighting. But it is still useful to make the framework more robust and more explainable.

\subsection{Unlearning Ability}\label{UnlearnAbility}
Following the same setting as in~\cite{Chen000HZ21}, we evaluate the unlearning ability of GUIDE using the state-of-the-art privacy attack against machine unlearning. We take the aggregated model of GUIDE as the unlearned model after processing 100 random unlearning requests. Using an enhanced membership inference attack~\cite{Chen000HZ21}, the attacker with  access to the original model and the unlearned model could determine whether a specific node is indeed removed from the unlearned model. The ratio of member and non-member is set to 1:1. As shown in Table~\ref{UnlearningAbility1}, the AUC of membership inference attack on GUIDE is close to $50\%$ (random guess), showing that GUIDE is enough to conduct machine unlearning with low privacy risk.    

\begin{table}[t]
\centering
\caption{AUC of membership inference attack on GUIDE(\%).}
\begin{tabular*}{\linewidth}{@{\extracolsep{\fill}}l;{1pt/1pt}ccc} 
\hline\hline \multicolumn{1}{l;{1pt/1pt}}{Dataset} & SAGE  & GAT      & GIN                 \\ 
\hline\hline
\multicolumn{1}{l;{1pt/1pt}}{Cora}    & 51.34$\pm$0.08 & 49.78$\pm$0.02 & 53.57$\pm$0.19  \\
\multicolumn{1}{l;{1pt/1pt}}{CiteSeer} & 53.36$\pm$0.10 & 50.97$\pm$0.12 & 50.70$\pm$0.08  \\
\multicolumn{1}{l;{1pt/1pt}}{DBLP}    & 53.34$\pm$0.07 & 51.22$\pm$0.19 & 55.83$\pm$0.7   \\
\multicolumn{1}{l;{1pt/1pt}}{CS}      & 50.34$\pm$0.14 & 51.27$\pm$0.14 & 48.09$\pm$0.14  \\
\hline\hline
\end{tabular*}
\label{UnlearningAbility1}
\end{table}

The study on the sensitivity of GUIDE to the number of shards is provided in Appendix~\ref{SensitiAppendix}.

\vspace{-0.1in}
\section{Conclusions}
In this work, we proposed the first general framework, GUIDE, for solving the inductive graph unlearning problem. 
Generally speaking, GUIDE consists of three components: guided graph
partition with fairness and balance, efficient subgraph repairing, and similarity-based aggregation. Due to its exceptional performance compared with the existing methods, we believe this work could serve as a cornerstone for future work on inductive graph unlearning tasks in production machine learning systems. 

Although GUIDE offers advantageous performance, it comes with additional memory cost due to its subgraph repair, making each subgraph larger than the original size. Furthermore, a generalization of "partition fairness" to unsupervised graph learning is needed for further applications.

\section*{Acknowledgments}

 Di Wang and Cheng-Long Wang were supported by BAS/1/1689-01-01, URF/1/4663-01-01, FCC/1/1976-49-01, RGC/3/4816-01-01, and REI/1/4811-10-01 of King Abdullah University of Science and Technology (KAUST) and KAUST-SDAIA Center of Excellence in Data Science and Artificial Intelligence. 

\bibliographystyle{plain}
\bibliography{camera-ready}

\clearpage
\appendix
\section{Related Work}\label{related_work}

\noindent
\textbf{Machine Unlearning.} Existing machine unlearning methods can be categorized into two classes according to their different assumptions. The first one redesigns the learning and unlearning process of the model, which is also called the exact unlearning method. Those methods~\cite{CaoY15,BourtouleCCJTZL21,BrophyL21} usually add an additional layer between the raw data layer and the model layer. By dividing the entire dataset into several disjoint subsets or generating different summaries in the middle layer, those methods could limit the influence of each sample on the whole model. When receiving an unlearning request, they only need to update the related parts, which has low computation cost. Another way to implement exact unlearning is to store the historical parameters of the model. When unlearning the trained model, \cite{UllahM0RA21} proposes to update the batch samples during training by first coupling the data to be forgotten with the historical batch. After checking whether the updated parameter is related to the new dataset, the algorithm can reject this update without retraining or accept it by retraining the model. 
 
However, in many cases, the model is already trained. There is no chance for the service provider to redesign the model's training process. We can only use the trained model and the entire data. This assumption is more restrictive than the above one. Under such assumption, many 
approximate unlearning methods can be adapted~\cite{GinartGVZ19,GuoGHM20,GuptaJNRSW21,IzzoSCZ21}. Those methods calibrate the parameters of the model trained on the full dataset to an approximation of the parameters of the model trained on the retained dataset. ~\cite{GuoGHM20} proposes an approximate retraining approach by taking a single step of Newton's method. \cite{KohL17} studies how to estimate the influence of a particular training point on the model's prediction. Both methods need to compute the hessian matrix, which is the main bottleneck. Guided by the Newton method and Influence method, some heuristic methods usually choose to design a mechanism for learning some noise so that the influence of the data to be forgotten to the model is as small as possible~\cite{GolatkarCVPR20,GolatkarECCV20,GravesNG21,0001IIM21}.

\noindent
\textbf{Federated SubGraph Learning (FSGL).} Since there are various settings of FSGL, here we focus on the one that is closest to our problems. Specifically, we assume that each client's local data is a subgraph of the entire graph. Depending on how the subgraphs are distributed in Federated Learning, there are different types of FSGL. A typical setting is there is no joint node between subgraphs held by different clients, but there may be cross-subgraph missing links between the nodes in different clients. The central server aims to learn a  GNN model. ~\cite{ZhangYLSY21} designs FedSage+ with a novel missing neighbor generator NeighborGen following with the corresponding local and federated training process. When the cross-client connections are known, ~\cite{ChenLMW22} proposes FedGraph using a novel cross-client graph convolution operation to compress the embeddings before sharing for privacy. Considering a personalization scenario where the  sizes of decentralized subgraphs can be very small, ~\cite{wu2022federated} presents FedPerGNN, which aims to collaboratively train GNN models from decentralized users' data by exploiting high-order interaction information in a privacy-preserving manner and expanding local subgraphs.  Besides the above settings, in the financial field, the subgraph held by one bank may contain an out-nodes set that multiple entities can own. ~\cite{YuanMWZW21} designs an ontology-based subgraph matching method for the query answering task over graph federation. To solve the decentralized learning problem over a network of nodes without a central server, ~\cite{abs-1901-11173} proposes a peer-to-peer federated learning algorithm on the graph. In this method, data owners share their local models with their neighbors to estimate the optimal global model.  However, the above-mentioned methods cannot be applied in the graph unlearning task where each shard cannot share parameter information with others during training.

\noindent
\textbf{Fair Clustering} The target of fair clustering is to find a partition that each protected class must have approximately equal representation in every cluster. ~\cite{Chierichetti0LV17} formulates the fair K-center and K-median clustering problems. It proposes efficient approximation algorithms to decompose the dataset into many good fairlets so that the traditional clustering algorithms can be applied. ~\cite{BackursIOSVW19} then designs a nearly linear time complexity version for the fairlet decomposition algorithm. To find the trade-off between clustering cost and fairness representation, ~\cite{abs-2105-14172} proposes a stochastic alternating balance K-means algorithm for fair clustering. Aiming to incorporate the fairness definition to graph data, ~\cite{KleindessnerSAM19} develops  both normalized and unnormalized spectral clustering with fairness constraints. ~\cite{BoseH19} introduces an adversarial framework to enforce the fairness constraint on graph embeddings. Some studies focus on individual fairness where any two similar individuals should receive similar algorithmic outcomes, such as ~\cite{MahabadiV20, KangHMT20}. However, the definition of individual fairness is totally different from the fairness discussed in this paper. The discussions on individual fairness are beyond our scope. To the best of our knowledge, our work is the first study on graph partition with both fairness and balance constraints.

\section{Spectral Clustering and Spectral Rotation Review}\label{spectral_scandsr}
Denote $\mathbf{W}\in \mathbb{R}^{n\times n}$ as the weighted adjacency matrix of an undirected graph $\mathcal{G=(V,E)}$ with $n$ nodes, where $w_{ij}$ corresponds to the weight of the edge between node $i$ and node $j$. We can calculate $\mathcal{G}$'s degree matrix $\mathbf{D}$ ($\mathbf{D}$ is diagonal) whose the $i$-th diagonal element $d_{i} = \sum_{j=1}^{n} W_{ij}$. Note that when $w_{ij}=1$ if node $i$ and $j$ are connected and $w_{ij}=0$ otherwise, the value of $d_{i}$ is just the number of neighboring nodes of node $i$. The standard graph partition problem with a fixed number of groups $v$ aims to find a partition of the graph such that the edges between different groups have low weights and the edges within a group have high weights. For each graph partition $\mathcal{V} = \dot{\cup}_{j\in [v]} \mathcal{V}_j$, 
one can encode it by  a {\bf group-membership indicator (binary) matrix} $\mathbf{Y}\in \mathbb{R}^{n\times v}$ and its normalized matrix $\mathbf{H}\in \mathbb{R}^{n\times v}$with  
\begin{equation}
H_{ij} = \left \{
\begin{aligned}
1/ \sqrt{|\mathcal{V}_j|},  & \quad i \in \mathcal{V}_j \\
0,  & \quad \text{otherwise}
\end{aligned}
\right. ,
Y_{ij} = \left \{
\begin{aligned}
1,  & \quad  i \in \mathcal{V}_j \\
0,  &\quad  \text{otherwise}
\end{aligned}
\right. .
\label{e0}
\end{equation}
We denote the set for all such $\mathbf{H}$ and $\mathbf{Y}$ as $\mathcal{H}$ and $\mathcal{Y}$ respectively. Denote the unnormalized Laplacian matrix as $L=W-D$, the RatioCut objective function can be written as $RatioCut(\mathcal{V}_1,\dots,\mathcal{V}_v) = Tr(\mathbf{H}^{\intercal}\mathbf{LH}) $ with $\mathbf{H}\in \mathcal{H}$~\cite{Luxburg07}. Spectral Clustering (SC) could be derived as a relaxation of such graph partition problem by replacing the $\mathbf{H}$ in eq.(\ref{e0}) with a relaxed constraint that  $\mathbf{H^{\intercal}H = I}$. Thus, the problem can be formulated as 
\begin{equation}
\begin{aligned}
    \min_{\mathbf{H}} &\quad Tr(\mathbf{H}^{\intercal}\mathbf{LH}), 
   \quad s.t. \quad \mathbf{H}^{\intercal}\mathbf{H} = \mathbf{I} 
\end{aligned}
\label{e1}
\end{equation}
The optimal solution of $\mathbf{H}$ is the top $v$ eigenvectors of the unnormalized Laplacian matrix $\mathbf{L}$. For the normalized Laplacian matrix $\mathbf{L}_{sym}=\mathbf{D}^{-1/2}\mathbf{LD}^{-1/2} = \mathbf{I} - \mathbf{D}^{-1/2}\mathbf{WD}^{-1/2}$,  its normalized SC problem is similar to (\ref{e1}),
where the Laplacian matrix $\mathbf{L}$ is replaced with $\mathbf{L}_{sym}$, and the constraint $\mathbf{H^{\intercal}H = I}$ is replaced by $\mathbf{H^{\intercal}DH = I}$. We refer the readers to the 
 survey~\cite{Luxburg07} for more details. 

However, due to the relaxed constraint, the optimal solution $\mathbf{H}$ in the above optimization problem cannot  directly imply a partition.  

Thus, we need to further apply a discretization procedure (such as  algorithms for K-Means clustering) on $\mathbf{H}$ to get the final partition. 

In summary, a standard SC algorithm can be conducted in three steps: (1) Calculate the Laplacian matrix $\mathbf{L}$; (2) Calculate the spectral embedding $\mathbf{H}\in \mathbb{R}^{n\times v} $ of this Laplacian matrix; (3) Apply some discretization procedure on $\mathbf{H}$ to get the final partition with $v$ groups.

Recently, researchers propose to further use spectral rotation on the embedding matrix for better clustering results than using  K-Means algorithms~\cite{HuangNH13a}. Generally speaking, for a given $\mathbf{H}$, the optimal indicator matrix $\mathbf{Y}$ can be obtained by solving the following problem:
\begin{equation}
\begin{aligned}
    \min_{\mathbf{Y}, \mathbf{R}  } &\quad  \lVert \mathbf{HR - Y}  \rVert_2^2, 
   \quad s.t. \quad \mathbf{R}^{\intercal}\mathbf{R} = \mathbf{I} 
\end{aligned}
\label{e2}
\end{equation}
where $\mathbf{R}\in \mathbb{R}^{v\times v}$ is an orthogonal matrix. This problem can be solved by an alternating minimization method.

\section{Graph Partition Algorithms}~\label{algorithmspart} 

Here we summarize the proposed two graph partition algorithms: GPFB-Fast in Algorithm~\ref{algo1} and GPFB-SR in Algorithm~\ref{algo2}. 

\RestyleAlgo{ruled}
\begin{algorithm}[h]
\LinesNumbered
\KwInput{The weighted adjacency matrix $\mathbf{W}$, degree matrix $\mathbf{D}$, balanced and fair guided matrix $\mathbf{\widetilde{M}}$, label-membership indicator matrix $\mathbf{F}$, number of subgroups $v$, regularization parameters $\alpha$.}
\KwOutput{New indicator matrix $\mathbf{Y}$.}

\textbf{Initialize}: $\mathbf{H,W,D}$.

Let $\mathbf{B}=\alpha \mathbf{F\widetilde{M}}$. 

\While{not converge}
{
    \While{$\mathbf{H}$ does not converge}
    {
        Calculate $\mathbf{P} = 2(\mathbf{W-D})\mathbf{H} - \alpha \mathbf{FF^{\intercal} H}+2\mathbf{B}$.
        
        Calculate the reduced SVD of $\mathbf{P}$ that $\mathbf{P=U}\Sigma \mathbf{V^{\intercal}}$.
        
        Update $\mathbf{H = UV^{\intercal}}$.
    }
    
    Applying K-means clustering to the rows of $\mathbf{H^{*}}$ to get $\mathbf{Y}$.
}
 
\caption{GPFB-Fast}
\label{algo1}
\end{algorithm}

\RestyleAlgo{ruled}
\begin{algorithm}[h]
\LinesNumbered
\KwInput{The weighted adjacency matrix $\mathbf{W}$, degree matrix $\mathbf{D}$, balanced and fair guided matrix $\mathbf{\widetilde{M}}$, fair guided matrix $\mathbf{F}$, number of subgroups $v$, regularization parameters $\alpha$, $\beta$.}
\KwOutput{New indicator matrix $\mathbf{Y}$.}

\textbf{Initialize}: $\mathbf{H,Y,R,W,D}$.

Let $\mathbf{B} = \alpha \mathbf{F\widetilde{M}}$.

\While{not converge}
{
    \textbf{Updating Rotation Matrix $\mathbf{R}$}: $\mathbf{R = UV^{\intercal}}$, where $\mathbf{H^{\intercal}}\mathbf{D}^{-\frac{1}{2}}\mathbf{Y(Y^{\intercal}DY)}^{-\frac{1}{2}}   = \mathbf{U}\Sigma \mathbf{V^{\intercal}} $.

    \textbf{Updating Embedding Matrix $\mathbf{H}$}: Let $\mathbf{A}= \mathbf{B} + \beta \mathbf{D}^{-\frac{1}{2}}\mathbf{Y(Y^{\intercal}DY)}^{-\frac{1}{2}} $. 
    
    \While{$\mathbf{H}$ does not converge}
    {
        Calculate $\mathbf{P} = 2(\mathbf{ W-D})\mathbf{H} - \alpha \mathbf{FF^{\intercal}H}+2\mathbf{A}$.
        
        Calculate the reduced SVD of $\mathbf{P}$ that $\mathbf{P}=\mathbf{U}\Sigma \mathbf{V^{\intercal}}$.
        
        Update $\mathbf{H = UV^{\intercal}}$.
    }
    
    \textbf{Updating Indicator Matrix $\mathbf{Y}$}: Let $\mathbf{a,b,c} \in \mathbb{R}^{v}$, $\mathbf{\widetilde{H}=HR}$.
    
    \While{$Y$ does not converge}
       {
        $\mathbf{a =1((\mathbf{D}^{-\frac{1}{2}}HR)\circ Y), b = 1Y^{\intercal}DY}$.
        
    	\For{$i=1$ to $n$}    
        { 
        	\For{$j=1$ to $v$}    
            { 
                $c_{j}=\frac{aj+\widetilde{h}_{ij}(1-y_{ij})}{\sqrt{b_{j}+1-y_{ij}}} - \frac{aj-\widetilde{h}_{ij}y_{ij}}{\sqrt{b_{j}-y_{ij}}} $.
            }
            
        $    
        y_{ij} =
            \begin{cases}
              1, & j= \text{arg} \max_{j\in[1,v]} c_{j}\\
              0. & \text{else}
            \end{cases} 
        $
        }
       }
}
 
\caption{GPFB-SR}
\label{algo2}
\end{algorithm}

\section{Omitted Proofs }\label{appendixproof}
\begin{proof}[{\bf Proof of Theorem \ref{theorem1}}]
By the defintion of $F$ and $Y$ we can easily see that  $\forall s\in [h], j\in [v]$,
$$
[\mathbf{F^{\intercal}Y}]_{s,j} =  |\mathcal{C}_s \cap \mathcal{V}_j |, 
$$  i.e., $[F^TT]_{s,j}$ is just the number of nodes with label $s$ in $\mathcal{V}_j$. Thus 
\begin{align*}
    [F^TT]_{s,j}=M_{s,j}=\frac{|\mathcal{C}_s|}{v}	\iff  |\mathcal{C}_s \cap \mathcal{V}_j |=\frac{|\mathcal{C}_s|}{v}.
\end{align*}
\end{proof}

\begin{proof}[{\bf Proof of Theorem \ref{theorem2}}]

If the partition is fair and balanced, then we have $|\mathcal{V}_j|=\frac{n}{v}$ and $|\mathcal{C}_s \cap \mathcal{V}_j |=\frac{|\mathcal{C}_s|}{v}$. Thus we have 
\begin{align*}
 &[\mathbf{F^{\intercal}H}]_{s,j}  = \frac{|\mathcal{C}_s \cap \mathcal{V}_j |}{\sqrt{|\mathcal{V}_j|}}=\frac{|\mathcal{C}_s|}{\sqrt{nv}}. 
\end{align*}
For the other side, suppose $[\mathbf{F^{\intercal}H}]_{s,j}  = \frac{|\mathcal{C}_s \cap \mathcal{V}_j |}{\sqrt{|\mathcal{V}_j|}}=\frac{|\mathcal{C}_s|}{\sqrt{nv}}$ then we have $\frac{|\mathcal{C}_s \cap \mathcal{V}_j |}{|\mathcal{C}_s|}=\frac{\sqrt{|\mathcal{V}_j|}}{\sqrt{nv}}$. That is 
\begin{align*}
    1=\sum_{j=1}^v\frac{|\mathcal{C}_s \cap \mathcal{V}_j |}{|\mathcal{C}_s|}=\frac{\sum_{j=1}^v\sqrt{|\mathcal{V}_j|}}{\sqrt{nv}}\implies  \sum_{j=1}^v\sqrt{|\mathcal{V}_j|}= \sqrt{nv}. 
\end{align*}
However, on the other hand, by Cauchy's inequality we have 
\begin{equation*}
    nv=(\sum_{j=1}^v\sqrt{|\mathcal{V}_j|})^2\leq v \sum_{j=1}^v \sqrt{|\mathcal{V}_j|}^2= v n, 
\end{equation*}
where the equality holds since $\{\mathcal{V}_j\}_{j=1}^r$ is a partition. That means the equality of Cauchy inequality  holds. Thus we have have $|\mathcal{V}_j|$ are the same for all $j\in [v]$, i.e., $|\mathcal{V}_j|=\frac{n}{v}$. Then we can get 
\begin{equation*}
[\mathbf{F^{\intercal}H}]_{s,j}  = \frac{|\mathcal{C}_s \cap \mathcal{V}_j |}{\sqrt{|\mathcal{V}_j|}}=\frac{|\mathcal{C}_s|}{\sqrt{nv}}\implies |\mathcal{C}_s \cap \mathcal{V}_j |=|\mathcal{C}_s|/v.
\end{equation*}
Thus the partition is both balanced and fair. 

\end{proof}
\subsection{Solving Problem (\ref{ae2_4})}\label{solvingalg2}

When $\mathbf{Y}$ and $\mathbf{H}$ is fixed, we can find the closed-form solution of $\mathbf{R}$ for problem (\ref{ae2_4}) directly. 
\begin{lemma}
For the following optimization problem 
\begin{equation}
    \min_{\mathbf{R^{\intercal}R=I}} \quad \lVert \mathbf{A-BR}  \rVert_2^2, 
\end{equation}
there is a closed-form solution of $\mathbf{R}$, that is $\mathbf{R^{*}=UV^{\intercal}}$, where $\mathbf{U,V}$ is constituted by the left and right singular vectors of $\mathbf{B^{\intercal}A}$ respectively~\cite{schonemann1966generalized}.
\label{lemma3}
\end{lemma}
Thus, $\mathbf{R}$ is updated according to 
\begin{equation}
    \mathbf{R}^{*} = \mathbf{UV^{\intercal}}, 
\label{e2_51}
\end{equation}
where $\mathbf{H^{\intercal}}\mathbf{D}^{-\frac{1}{2}}\mathbf{Y(Y^{\intercal}DY)}^{-\frac{1}{2}}  = \mathbf{U}\Sigma \mathbf{V^{\intercal}} $.

When $\mathbf{Y}$ and $\mathbf{R}$ is fixed, problem (\ref{ae2_4}) is equal to 
\begin{equation}
\begin{aligned}
    \max_{\mathbf{H}} &\quad  Tr(\mathbf{H}^{\intercal}(\mathbf{W - D} - \alpha \mathbf{FF^{\intercal})H} \\
    & + 2\mathbf{H^{\intercal}}(\alpha \mathbf{F\widetilde{M}} + \beta \mathbf{D}^{-\frac{1}{2}}\mathbf{Y(Y^{\intercal}DY)}^{-\frac{1}{2}} ) ).\\
    s.t. &\quad \mathbf{H^{\intercal}H = I}
\end{aligned}
\label{e2_52}
\end{equation}
This quadratic problem over the Stiefel manifold can be also solved similarly with Algorithm \ref{algo1}.

When $\mathbf{H}$ and $\mathbf{R}$ are fixed, it is not hard to write the problem (\ref{ae2_4}) as  
\begin{equation}
    \max_{\mathbf{Y}} Tr(\mathbf{R^{\intercal}H^{\intercal}\mathbf{D}^{-\frac{1}{2}}Y(Y^{\intercal}DY)}^{-\frac{1}{2}}). 
\label{e2_7}
\end{equation}
This problem can be solved iteratively.

\section{Subgraph Repairing Strategies}~\label{SubgraphRepairing}
\begin{figure}[th]
\centering
    \includegraphics[width=0.45\textwidth]{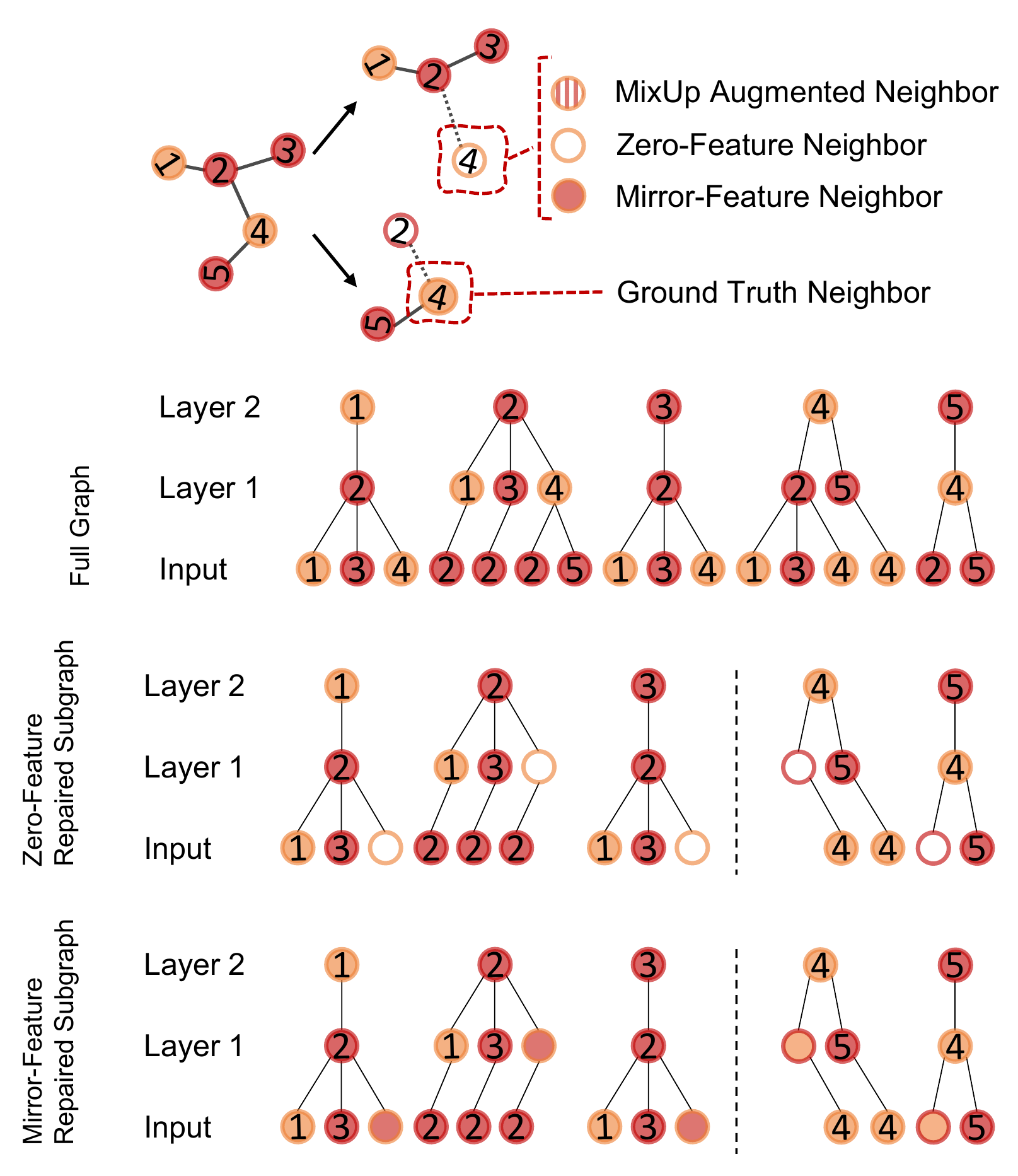}
\caption{Computation graphs of a 2-layer GNN on the full training graph, Zero-Feature repaired subgraphs, and Mirror-Feature repaired subgraphs.}
\label{Picture5}
\end{figure}

Figure \ref{Picture5} illustrates the differences between the computation graphs of a 2-layer GNN on the full training graph and the repaired subgraphs. Compared to Figure \ref{Picture3}, we can see that the Zero-Feature and Mirror-Feature repairing strategies are helpful in repairing GNNs' computation graphs.

\section{Details of Experimental Settings}

\subsection{Datasets}\label{otherdetaildata}

\begin{table}[h]
\caption{Dataset statistics}
\begin{center}
\begin{tabular*}{\columnwidth}{p{1cm}p{1cm}p{1cm}cp{1cm}p{1cm}}
\hline
Dataset  & Type     & \#Node & \#Edge & \#Feature & \#Label \\ \hline
Cora     & Citation & 2,708   & 10,556   & 1,433 & 7  \\
CiteSeer & Citation & 3,327   & 9,104   & 3,703 & 6  \\
DBLP  & Citation & 17,716  & 105,734 & 1,639 & 4  \\
CS       & Coauther & 18,333  & 163,788 & 6,805 & 15 \\  
Elliptic  & Payment & 203,769  &234,355 & 165 & 2 \\  
 \hline
\end{tabular*}    
\end{center}
\label{datatable}
\end{table}

The statistical information of all datasets is summarized in Table~\ref{datatable}. The citation network datasets Cora and CiteSeer are from ~\cite{YangCS16}. The citation network dataset DBLP is from~\cite{BojchevskiG18}. Nodes represent documents and edges represent citation links. The coauthor network dataset CS is from\cite{abs-1811-05868} whose nodes represent authors that are connected by an edge if they co-authored a paper. Given paper keywords for each author's papers, the task is to map authors to their respective fields of study. 

\subsection{Baseline Setup}\label{otherdetailbaseline}

\begin{itemize}
    \item Scratch: All GNN models are trained on the entire training graph without partitioning. When receiving an unlearning request, the Scratch method will retrain the GNN models from the new training graph, which directly removes all the requested nodes and edges from the original graph. 
    
    \item Random: In this method, the training graph is randomly partitioned based on node IDs. The prediction of this method on the test graph is the average aggregation of all submodels' predictions. After receiving an unlearning request, this method only retrains the corresponding subgraphs and their submodels.
    
    \item Eraser-BLPA: A fast method of GraphEraser with the training graph is partitioned by BLPA~\cite{abs-2103-14991}\footnote{https://github.com/MinChen00/Graph-Unlearning}. The max iteration of BLPA is set to 100. The optimal importance scores of all submodels are learned by LBAggr on the node embedding vectors from the training graph. After receiving an unlearning request, Eraser-BLPA updates the corresponding subgraphs and submodels. Although ~\cite{abs-2103-14991} empirically shows that they can use only a small random subset of nodes from the training graph to learn the optimal importance scores, here we train LBAggr using all node embeddings to achieve its best performance in the inductive setting. When  there is a node-unlearning request, we retrain LBAggr to update the optimal importance scores. 
  
    \item Eraser-BEKM: Another method of GraphEraser~\cite{abs-2103-14991}. A pre-trained GNN model is applied to get the node embeddings of the training graph. The training graph is then partitioned by BEKM based on the node embeddings~\cite{abs-2103-14991}. We generate the node embeddings with the GraphSAGE model for each training graph to achieve its best performance in the inductive setting. The embedding dimension is set to 256. Other settings remain the same as in Eraser-BLPA.
    
\end{itemize}

\textbf{GNN Models.} All GNN models are implemented following the setting of Pytorch Geometric Library\footnote{https://github.com/pyg-team/pytorch\_geometric}. Among them, all GNN models except APPNP have 2 GNN layers. APPNP consists of 2 linear layers and 1 propagation layer. We use the Adam optimizer with a learning rate of 0.01 and weight decay of 0.0005. We also report the performance of SGC~\cite{WuSZFYW19} in the inductive setting, which is a simplified GCN model and has been applied in the Certified Graph Unlearning method~\cite{abs-2206-09140}. We train SGC for 200 epochs using Adam with a learning rate of 0.1. All GNN models are trained with 200 epochs.

\section{Additional Experimental Results}

\subsection{Model Utility with SGC and MLP}\label{SGDAppendix}

\begin{table*}[ht]
\centering
\caption{Node classification accuracy of SGC on four datasets(\%).}
\begin{tabular*}{\linewidth}{@{\extracolsep{\fill}}ll;{1pt/1pt}ccc;{1pt/1pt}cc} 
\hline\hline 
\multicolumn{1}{l;{1pt/1pt}}{DataSet}        & Scratch                         & Random                          & Eraser-BLPA                            & Eraser-BEKM                            & GUIDE-Fast                       & GUIDE-SR                    \\ 
\hline\hline
\multicolumn{1}{l;{1pt/1pt}}{Cora}            & 38.52$\pm$0.01  &33.84$\pm$0.06   &31.66$\pm$0.05   &33.04$\pm$0.14   &33.27$\pm$0.06   &33.40$\pm$0.05    \\ 
\multicolumn{1}{l;{1pt/1pt}}{CiteSeer}       & 57.05$\pm$0.16  &52.74$\pm$0.29   &41.35$\pm$1.29   &39.23$\pm$1.61   &51.37$\pm$0.30   &49.17$\pm$0.58 \\ 
\multicolumn{1}{l;{1pt/1pt}}{DBLP}             & 70.47$\pm$0.00  &59.85$\pm$0.01   &70.06$\pm$0.02   &68.91$\pm$0.03   &69.41$\pm$0.02   &64.54$\pm$0.01  \\
\multicolumn{1}{l;{1pt/1pt}}{CS}           & 22.77$\pm$0.00  &23.18$\pm$0.01   &22.86$\pm$0.01   &22.80$\pm$0.01   &23.05$\pm$0.01   &23.06$\pm$0.01 \\
\hdashline[1pt/1pt]
\multicolumn{1}{l;{1pt/1pt}}{Normalized Score}  & 100.00 & 0.00 &-28.68	&-36.31	&36.51	&12.98 \\
\hline\hline
\end{tabular*}
\label{SGCmodel}
\end{table*}

\begin{table*}[th]
\centering
\caption{Node classification accuracy of MLP on four datasets(\%).}
\begin{tabular*}{\linewidth}{@{\extracolsep{\fill}}ll;{1pt/1pt}ccc;{1pt/1pt}cc} 
\hline\hline 
\multicolumn{1}{l;{1pt/1pt}}{DataSet}        & Scratch                         & Random                          & Eraser-BLPA                            & Eraser-BEKM                            & GUIDE-Fast                       & GUIDE-SR                    \\ 
\hline\hline
\multicolumn{1}{l;{1pt/1pt}}{Cora}            & 87.46$\pm$0.00 & 52.50$\pm$0.17          & 40.65$\pm$0.07          & 48.41$\pm$0.12          & 52.31$\pm$0.10          & 52.85$\pm$0.14  \\ 
\multicolumn{1}{l;{1pt/1pt}}{CiteSeer}      & 80.29$\pm$0.00 & 68.57$\pm$0.03 & 62.59$\pm$0.27          & 66.59$\pm$0.08          & 68.29$\pm$0.03          & 68.00$\pm$0.03           \\ 
\multicolumn{1}{l;{1pt/1pt}}{DBLP}           & 80.16$\pm$0.00 & 65.79$\pm$0.01          & 65.93$\pm$0.01          & 67.43$\pm$0.00 & 67.00$\pm$0.01          & 65.67$\pm$0.01           \\
\multicolumn{1}{l;{1pt/1pt}}{CS}             & 78.08$\pm$0.02 & 69.74$\pm$0.01          & 55.02$\pm$0.02          & 65.57$\pm$0.03          & 69.99$\pm$0.01          & 70.16$\pm$0.01  \\
\hdashline[1pt/1pt]
\multicolumn{1}{l;{1pt/1pt}}{Normalized Score} & 100.00 & 0.00 &-65.11	&-16.80	&2.12	&0.08 \\
\hline\hline
\end{tabular*}
\label{MLPutility}
\end{table*}

To compare with the Certified Graph Unlearning method~\cite{abs-2206-09140}, we adapt the SGC model to the Scratch and the above five graph unlearning methods. As shown in Table~\ref{SGCmodel}, the performance of the SGC model is inconsistent across the four datasets. The inductive node classification accuracy of SGC on the CS dataset is only $22.77\%$. The Random method even achieves higher performance than the Scratch method. This means that SGC failed to apply its learned structural information from the full training graph to generate a good representation of the test graph on the CS dataset. Thus, when calculating the normalized score, we ignore the abnormal case of the CS dataset. Compared with all normalized scores, we can see GraphEraser behaves worse than the Random method while GUIDE still maintains its superiority.

We also report the results of a 2-layer MLP (Multi-Layer Perceptron) model without considering the graph structure in Table~\ref{MLPutility}. In this case,  it is  reasonable that GUIDE achieves comparable results to Random as it could provide a partition with a similar fairness score compared with  Random. However, GraphEraser cannot provide a fair partition, so its results are worse than Random. The comparable results (compared with Random) of GUIDE combined with MLP indicate that the fair graph partition produced by GUIDE can also be significant for future extensions using MLP. For example, a recent work~\cite{ZhangLSS22} has shown that the performance of MLPs can be significantly improved with GNN Knowledge Distillation. Furthermore, MLPs infer much faster than GNNs. 

\subsection{Sensitivity Analysis}\label{SensitiAppendix}
\begin{figure}[t]
\centering
\includegraphics[width=0.45\textwidth]{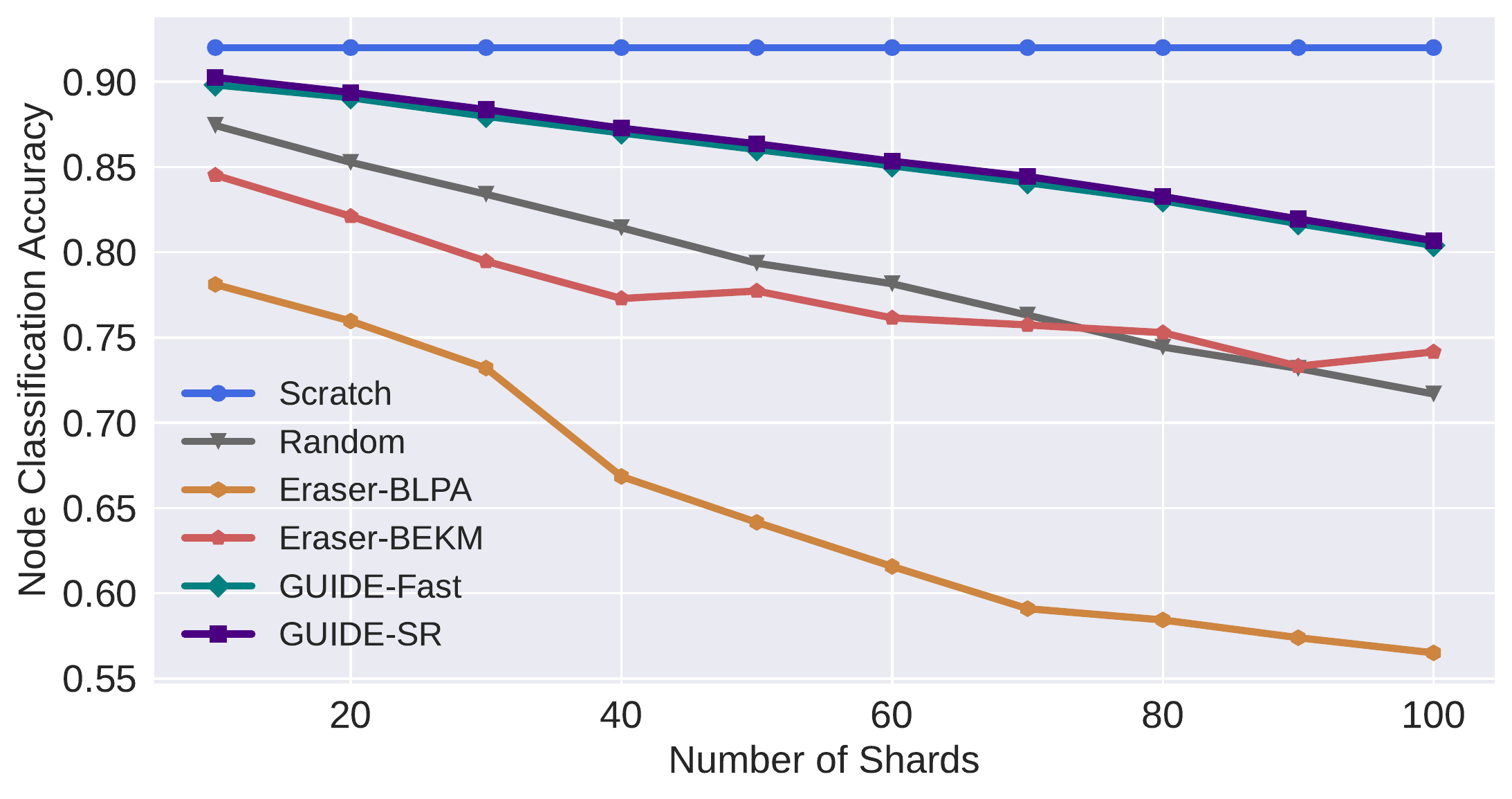}
\caption{Performance of different methods on CS. The curve of GUIDE-Fast overlaps with the curve of GUIDE-SR.}
\label{Picture7}
\end{figure}

\begin{figure}[h]
\centering
    \includegraphics[width=0.45\textwidth]{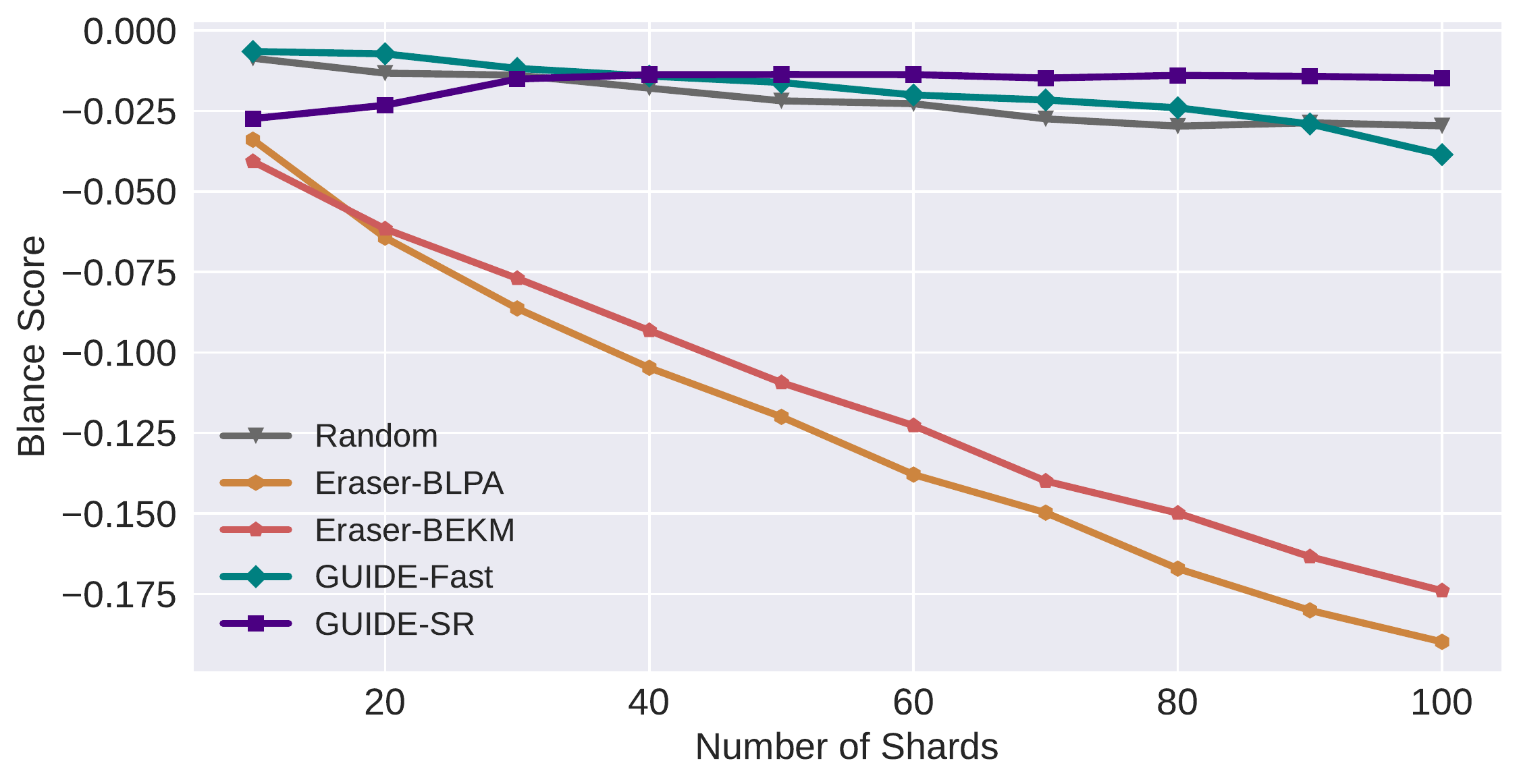}
\caption{Balance scores of different methods on CS.}
\label{Picture8}
\end{figure}

\begin{figure}[h]
\centering
    \includegraphics[width=0.45\textwidth]{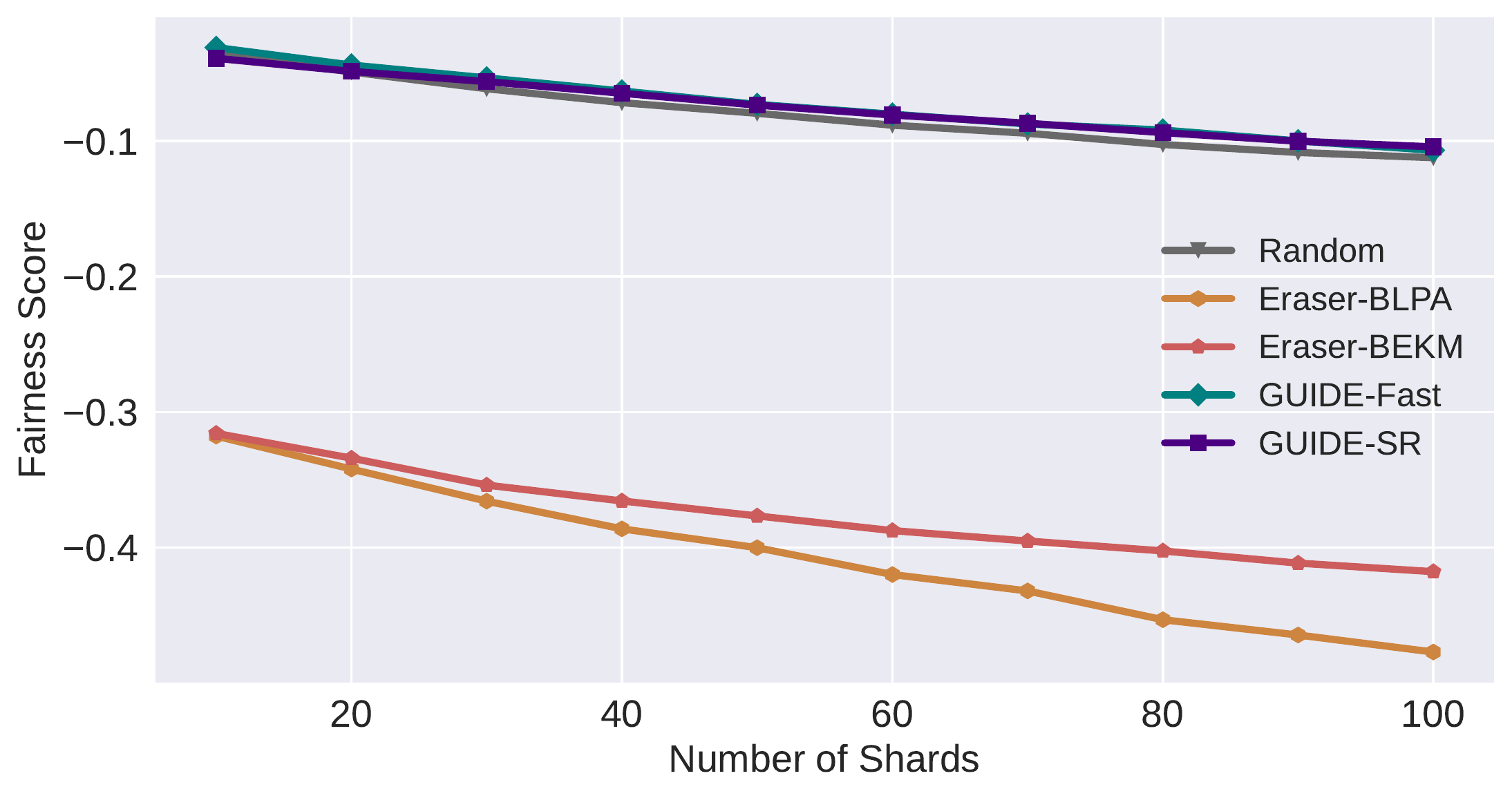}
\caption{Fairness scores of different methods on CS.}
\label{Picture9}
\end{figure}

\begin{figure}[h]
\centering
\begin{subfigure}{0.45\textwidth}
 \centering
 \includegraphics[width=\textwidth]{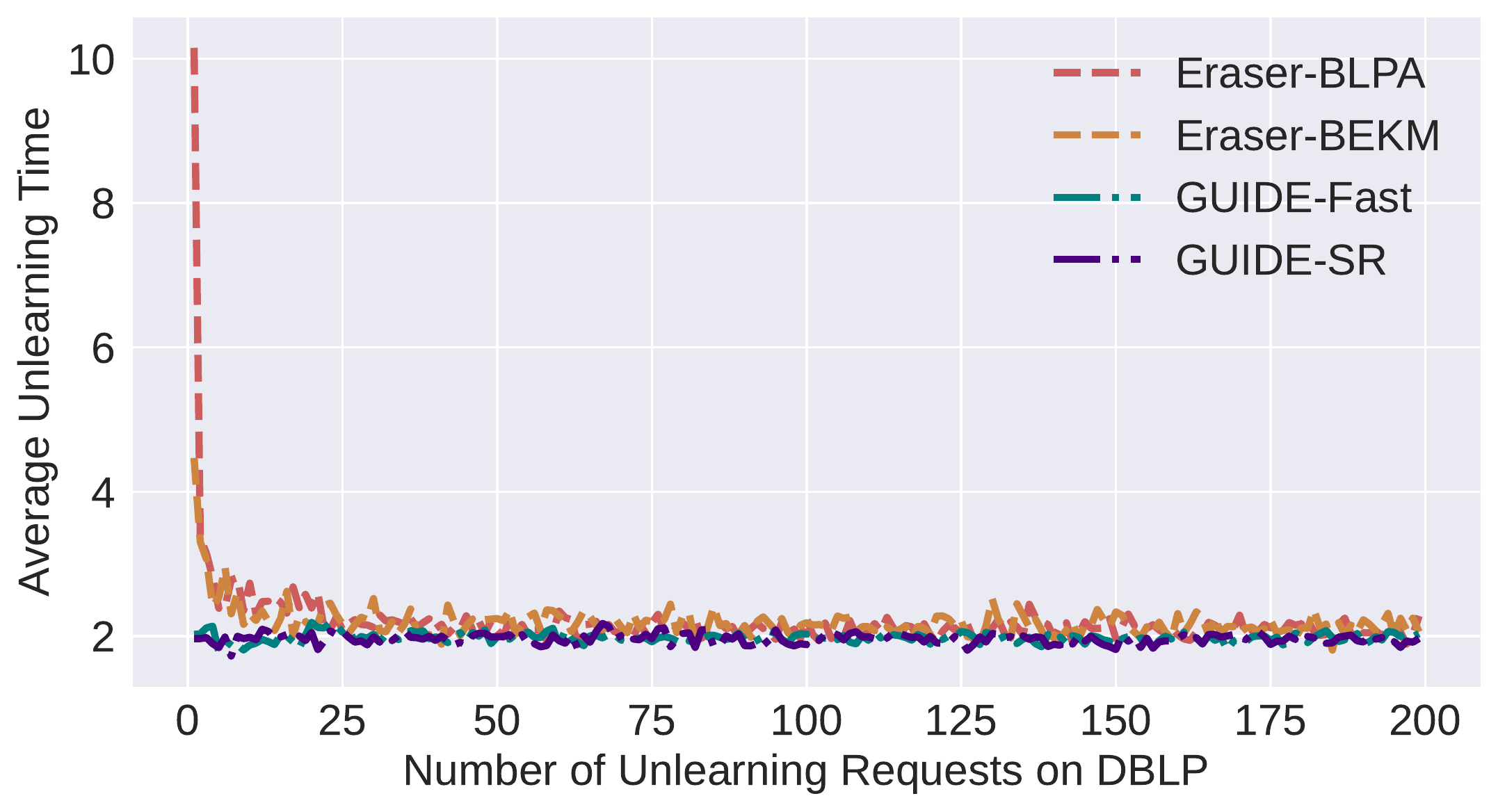}
\end{subfigure}
\hfill
\begin{subfigure}{0.45\textwidth}
 \centering
  \includegraphics[width=\textwidth]{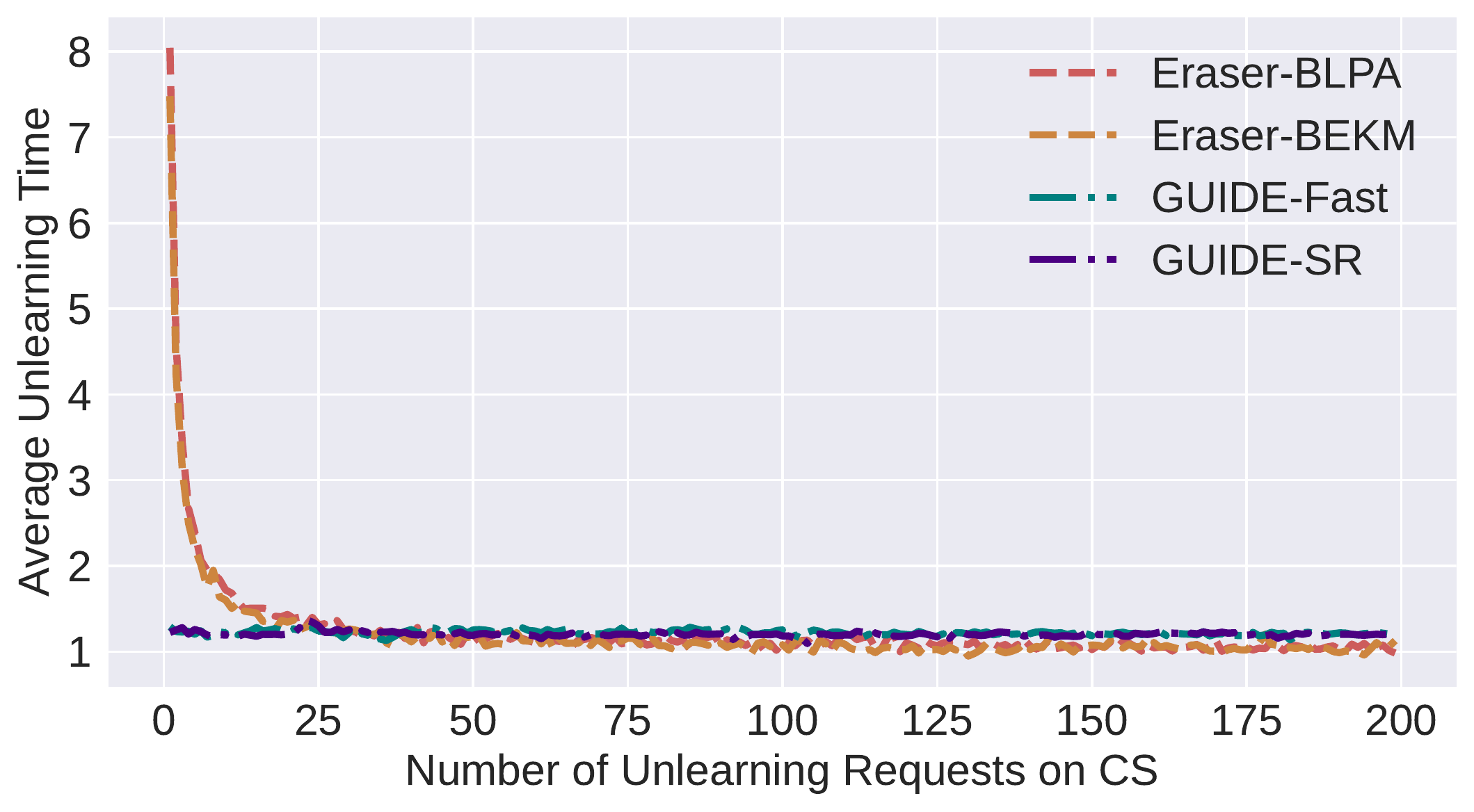}
\end{subfigure}
\hfill
\begin{subfigure}{0.45\textwidth}
 \centering
  \includegraphics[width=\textwidth]{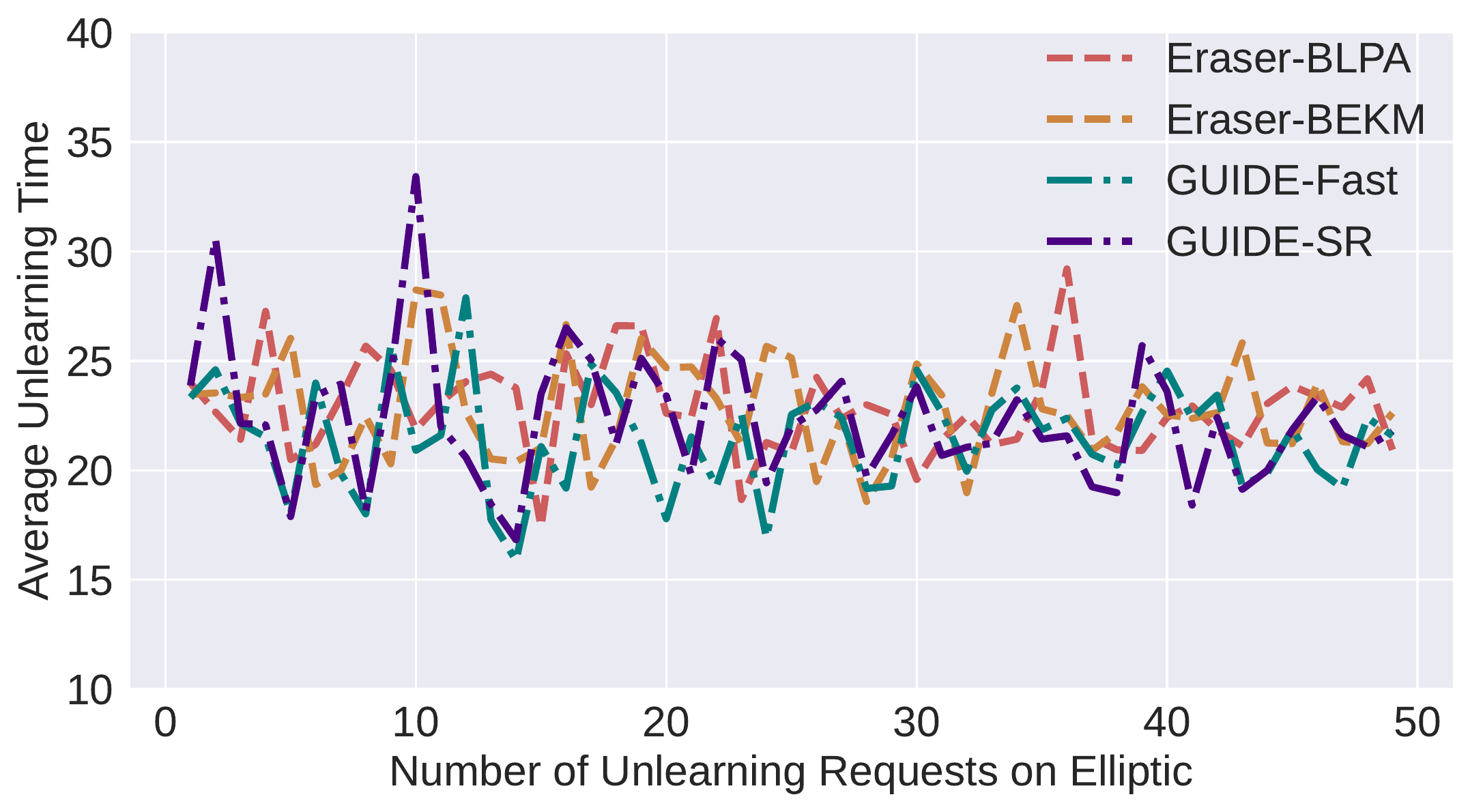}
\end{subfigure}
\caption{Average unlearning time of each involved shard.}
\label{PictureBatchUnlearningAve}
\vspace{-0.1in}
\end{figure}

\begin{figure*}[th]
\centering
\begin{subfigure}{0.49\textwidth}
 \centering
   \includegraphics[width=\textwidth]{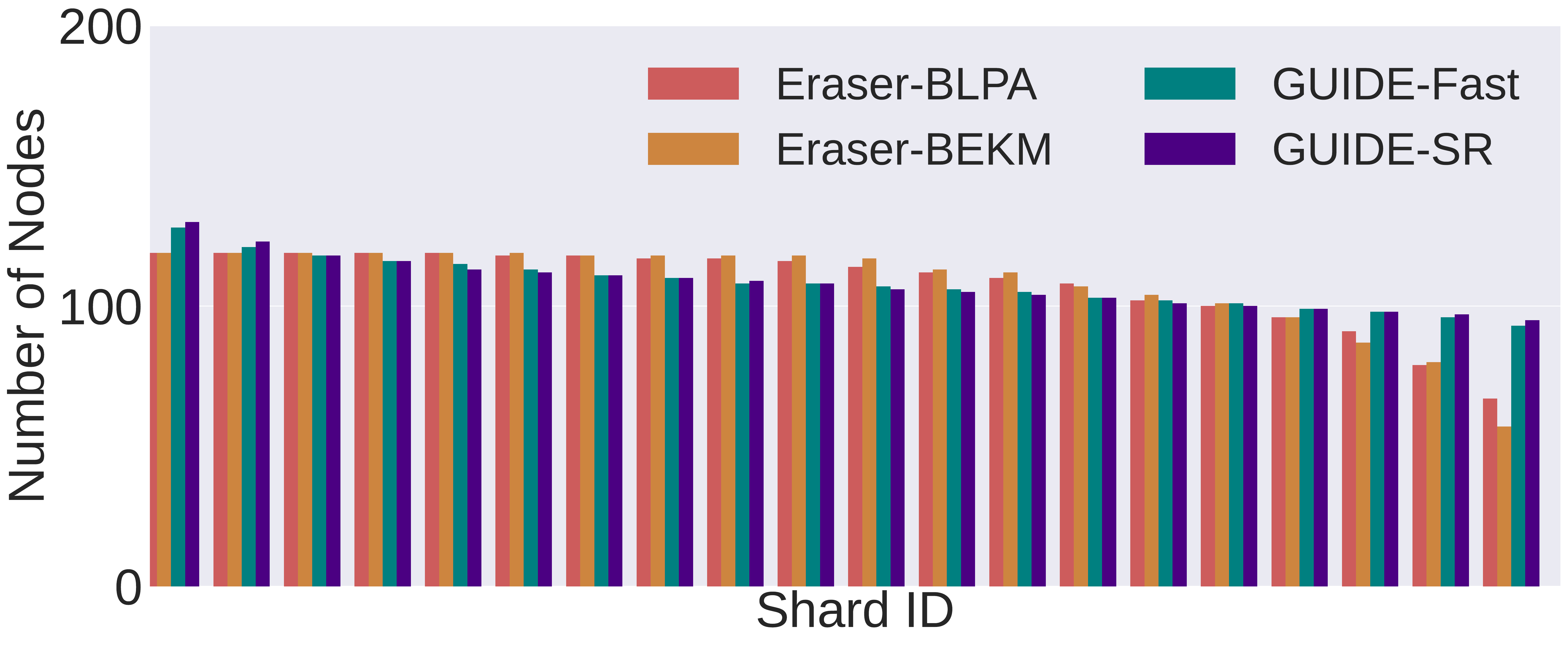}
 \subcaption{Cora}
\end{subfigure}
\hfill
\begin{subfigure}{0.49\textwidth}
 \centering
   \includegraphics[width=\textwidth]{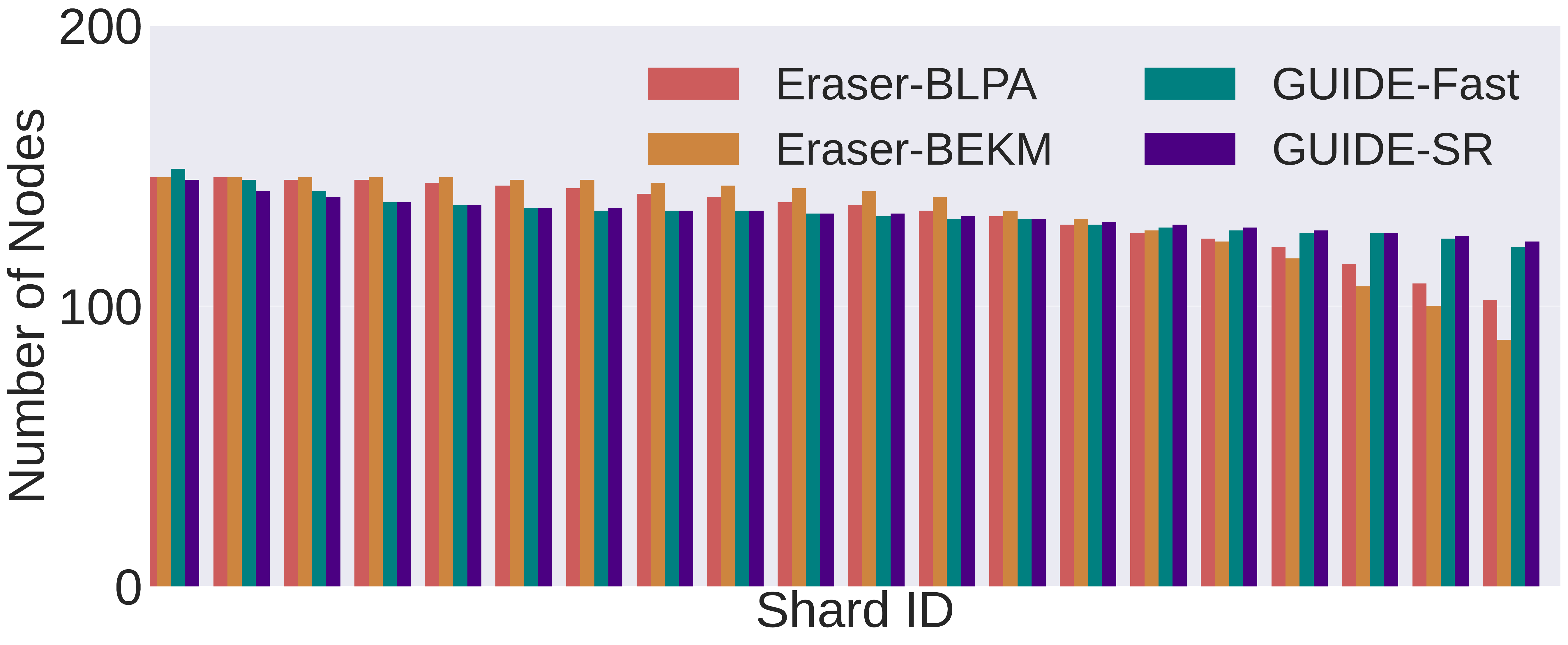}
 \subcaption{CiteSeer}
\end{subfigure}
\hfill
\begin{subfigure}{\textwidth}
 \centering
  \includegraphics[width=\textwidth]{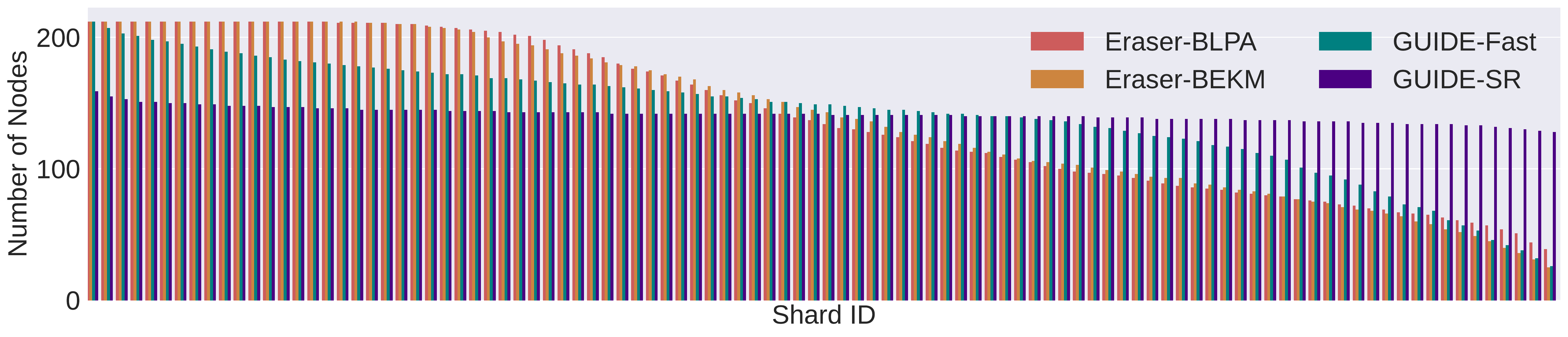}
 \subcaption{DBLP}
\end{subfigure}
\hfill
\begin{subfigure}{\textwidth}
 \centering
 \includegraphics[width=\textwidth]{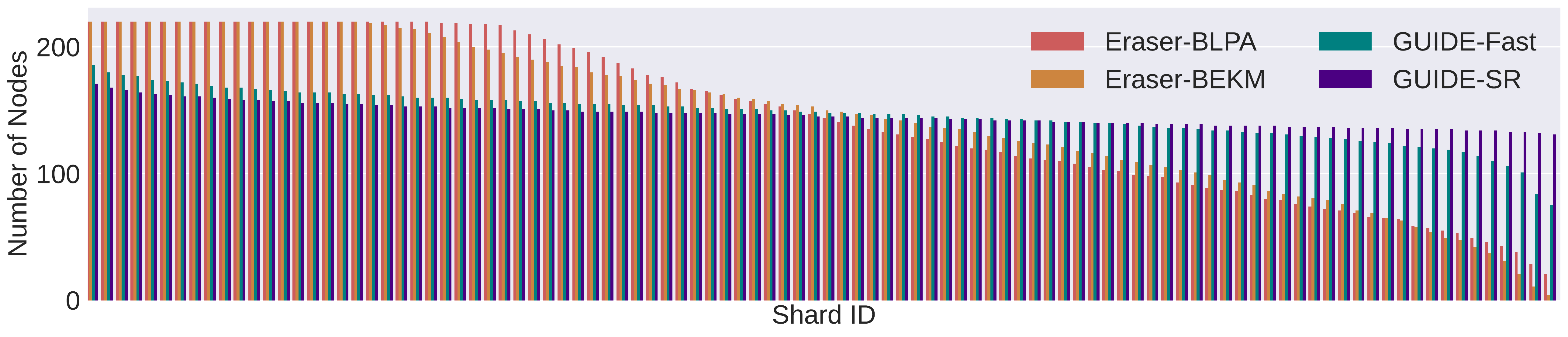}
 \subcaption{CS}
\end{subfigure}
\caption{Distributions of shard sizes partitioned by different methods on four datasets.}
\label{Picture0_}
\end{figure*}

\noindent
\textbf{Impact of Shard Number.} We study the sensitivity of GUIDE to the number of shards. The experiments are conducted on the CS dataset. The number of shards varies from 10 to 100. As shown in Figure~\ref{Picture7}, both GUIDE-Fast and GUIDE-SR show the same trend. When the number of shards is small (such as 10), the gap between GUIDE and Scratch is also small. As the number of shards becomes larger, the performance gap between GUIDE and scratch will gradually increase. This is reasonable as more shards lead to more loss of structural information. The performance of GUIDE-Fast and GUIDE-SR degrades more slowly than other methods. The reason is that our subgraph repair strategies reduce some side-effects of the partition on performance. 

As shown in Figure~\ref{Picture8}, the balance scores of GUIDE-Fast and GUIDE-SR keep a slight change similar to Random, while the balance scores of GraphEraser degrade faster than GUIDE. In Figure~\ref{Picture9}, we can see the fairness scores of all five methods show the same trend when the shard number becomes larger. Moreover, our two variants of GUIDE achieve almost the same fairness score as Random, which is much higher than GraphEraser. 

\subsection{Distribution of the Partitioned Shard Sizes}\label{PartitionSize}

We present the distributions of shard sizes partitioned by different methods on four datasets in Figure~\ref{Picture0_}. It is not hard to see that the graph partitions provided by GUIDE achieve more balanced distributions, especially on large datasets, i.e., DBLP and CS.

\subsection{Average Unlearning Time of Each Shard}\label{AveBatchUnlearning}

Figure~\ref{PictureBatchUnlearningAve} shows the average unlearning time of each shard involved. When the unlearning batch size is small, where the aggregate step dominates the total computation cost, it's easy to see that GUIDE has less time cost. When the unlearning batch size is large, the training time cost for submodels that need to be updated is the major part. Thus, we can see that the average unlearning time cost of each shard in GraphEraser is close to it in GUIDE. For Elliptic, since each model is trained with evolving graphs, the submodel training time always dominates the unlearning cost.

\subsection{Experiments for Transduction Tasks}\label{TransductiveAppendix}
The performance of all baselines and GUIDE in the transductive setting is reported in Table~\ref{TransductiveACC}. We can see that GUIDE has comparable performance to GraphEraser and Random, where all of them have large gaps compared with Scratch. Moreover, the performance gap between Scratch and other graph unlearning methods seems larger than it reported in \cite{abs-2103-14991}. There are many reasons for this. For example, in our experiments, we calculate the average results of all models on 10 random training/testing splits, which indicates that  the performance of each model can vary widely between different training/testing splits. While \cite{abs-2103-14991} reports the average results of many runs on a fixed split, i.e, their setting is different from ours. Besides, we use the same parameters as in the inductive setting as the experimental control. Increasing the width of the network embedding layer may improve performance, but it does not affect the experimental conclusions that rely on the relative results.

\begin{table*}[th]
\centering
\caption{Node classification accuracy of difference graph unlearning methods in the transductive setting(\%).}
\begin{tabular}{ll;{1pt/1pt}c;{1pt/1pt}c;{1pt/1pt}cc;{1pt/1pt}cc} 
\hline\hline
Dataset                   & Model      & Scratch                              & Random          & Eraser-BLPA     & Eraser-BEKM     & GUIDE-Fast      & GUIDE-SR         \\ 
\hline\hline
\multirow{4}{*}{Cora}     & SAGE       & $89.26\pm 0.02$                      & $59.17\pm 0.10$ & $45.82\pm 0.24$ & $54.81\pm 0.24$ & $49.72\pm 0.07$ & $51.61\pm 0.15$  \\
                          & GIN        & $84.16\pm 0.05$                      & $38.93\pm 0.10$ & $34.05\pm 0.05$ & $36.77\pm 0.09$ & $36.40\pm 0.06$ & $37.32\pm 0.09$  \\
                          & GAT        & $86.75\pm 0.03$                      & $33.86\pm 0.07$ & $32.64\pm 0.03$ & $33.35\pm 0.03$ & $36.82\pm 0.12$ & $39.59\pm 0.26$  \\
                          & GCN        & $89.32\pm 0.01$                      & $56.73\pm 0.15$ & $43.99\pm 0.13$ & $50.02\pm 0.07$ & $59.57\pm 0.14$ & $63.23\pm 0.07$  \\ 
\hdashline[1pt/1pt]
\multirow{4}{*}{CiteSeer} & SAGE       & $76.72\pm 0.01$                      & $68.83\pm 0.02$ & $65.56\pm 0.02$ & $67.25\pm 0.02$ & $66.29\pm 0.04$ & $65.94\pm 0.02$  \\
                          & GIN        & $70.38\pm 0.05$                      & $42.84\pm 0.09$ & $37.74\pm 0.04$ & $40.89\pm 0.12$ & $43.76\pm 0.14$ & $44.86\pm 0.28$  \\
                          & GAT        & $73.76\pm 0.02$                      & $43.44\pm 0.08$ & $36.96\pm 0.23$ & $35.32\pm 0.07$ & $48.39\pm 0.13$ & $49.76\pm 0.11$  \\
                          & GCN        & $77.16\pm 0.02$                      & $67.23\pm 0.02$ & $61.62\pm 0.07$ & $64.84\pm 0.02$ & $67.68\pm 0.02$ & $68.23\pm 0.01$  \\ 
\hdashline[1pt/1pt]
\multirow{4}{*}{DBLP}     & SAGE       & $85.53\pm 0.00$                      & $67.70\pm 0.06$ & $66.81\pm 0.01$ & $68.22\pm 0.05$ & $72.20\pm 0.01$ & $72.29\pm 0.01$  \\
                          & GIN        & $82.92\pm 0.01$                      & $59.77\pm 0.22$ & $65.08\pm 0.02$ & $63.78\pm 0.15$ & $40.52\pm 0.19$ & $44.99\pm 0.27$  \\
                          & GAT        & $82.89\pm 0.00$                      & $58.35\pm 0.24$ & $61.58\pm 0.12$ & $62.50\pm 0.11$ & $68.35\pm 0.09$ & $69.60\pm 0.02$  \\
                          & GCN        & $85.47\pm 0.00$                      & $74.72\pm 0.00$ & $70.02\pm 0.04$ & $72.48\pm 0.02$ & $72.83\pm 0.05$ & $75.49\pm 0.01$  \\ 
\hdashline[1pt/1pt]
\multirow{4}{*}{CS}       & SAGE       & $93.54\pm 0.00$                      & $75.66\pm 0.01$ & $57.11\pm 0.06$ & $74.33\pm 0.07$ & $64.93\pm 0.02$ & $67.30\pm 0.01$  \\
                          & GIN        & $84.60\pm 1.28$                      & $53.52\pm 0.09$ & $44.44\pm 0.12$ & $50.41\pm 0.27$ & $50.51\pm 0.05$ & $50.17\pm 0.02$  \\
                          & GAT        & $87.33\pm 0.00$                      & $37.36\pm 0.08$ & $45.38\pm 0.07$ & $43.59\pm 0.22$ & $45.91\pm 0.06$ & $50.69\pm 0.02$  \\
                          & GCN        & $88.77\pm 0.00$                      & $71.03\pm 0.01$ & $53.67\pm 0.06$ & $61.67\pm 0.27$ & $68.23\pm 0.01$ & $69.89\pm 0.01$  \\ 
\hdashline[1pt/1pt]
\multicolumn{2}{l;{1pt/1pt}}{Normalized Score}   & 100.00 & 0.00            & -20.18          & -6.73           & -3.97           & 2.75             \\
\hline\hline
\end{tabular}
\label{TransductiveACC}
\end{table*}

\end{document}